\documentclass[letter,11pt]{article}
\usepackage{fullpage}

\usepackage[cmex10]{amsmath}
\usepackage{epsfig,epsf,psfrag,amssymb,amsfonts,latexsym,graphicx,bm,cite,xcolor,url,array}
\usepackage[caption=false]{subfig}
\usepackage{fixltx2e}
\usepackage{cases}
\usepackage{verbatim}
\usepackage[mathscr]{eucal}
\usepackage{algorithm, algorithmic}
\usepackage[fleqn,tbtags]{mathtools}
\usepackage{amsthm}
\usepackage[fleqn,tbtags]{mathtools}

\usepackage[parfill]{parskip}

\usepackage{appendix}
\usepackage{cite}

\newtheorem{lem}{Lemma}

 \DeclareMathOperator{\off}{off}

\DeclareMathOperator{\Supp}{Supp}
 
 \DeclareMathOperator*{\argmin}{arg\,min}
 \DeclareMathOperator*{\argmax}{arg\,max}

\DeclareMathOperator{\hSigma}{\widehat{\Sigma}}

\newcommand\indep{\protect\mathpalette{\protect\independenT}{\perp}}
\def\independenT#1#2{\mathrel{\rlap{$#1#2$}\mkern2mu{#1#2}}}

\DeclarePairedDelimiter\norm{\lVert}{\rVert}

\def\lnorm{{\lvert\!\lvert\!\lvert}}
\def\rnorm{{\rvert\!\rvert\!\rvert}}
\DeclarePairedDelimiter\gennorm{\lnorm}{\rnorm}

 \def\0{{\bf 0}}

\def\viz{{viz.,\ \/}}

\def\qed{\hfill\hbox{${\vcenter{\vbox{
    \hrule height 0.4pt\hbox{\vrule width 0.4pt height 6pt
    \kern5pt\vrule width 0.4pt}\hrule height 0.4pt}}}$}}

\def\tcr{\textcolor{red}}

\definecolor{myred}{rgb}{0.3,0.0,0.7}
\definecolor{dkg}{rgb}{0.1,0.7,0.2}
\definecolor{dkb}{rgb}{0.0,0.2,0.8}

\def\tcdkb{\textcolor{dkb}}

\def\hJ{\widehat{J}}
 
\def\bfh{{\mathbf h}}

\def\bfx{{\mathbf x}}

\def\bfX{{\mathbf X}}

\def\mubf{\hbox{\boldmath$\mu$\unboldmath}}

\def\Nc{{\cal N}}

\def\Ebb{{\mathbb E}}

\def\tilSigma{{\widetilde{\Sigma}}}

\newcommand{\bprf}{\begin{myproof}}
\newcommand{\eprf}{\end{myproof}}
\newcommand{\bp}{\begin{psfrags}}
\newcommand{\ep}{\end{psfrags}}
\newcommand{\bl}{\begin{lemma}}
\newcommand{\el}{\end{lemma}}
\newcommand{\bt}{\begin{theorem}}
\newcommand{\et}{\end{theorem}}
\newcommand{\bc}{\begin{center}}
\newcommand{\ec}{\end{center}}
\newcommand{\bi}{\begin{itemize}}
\newcommand{\ei}{\end{itemize}}
\newcommand{\ben}{\begin{enumerate}}
\newcommand{\een}{\end{enumerate}}
\newcommand{\bd}{\begin{definition}}
\newcommand{\ed}{\end{definition}}
\def\beq{\begin{equation}}
\def\eeq{\end{equation}\noindent}
\def\beqn{\begin{eqnarray}}
\def\eeqn{\end{eqnarray} \noindent}
\def\beqnn{  \begin{eqnarray*}}
\def\eeqnn{\end{eqnarray*}  \noindent}
\def\bcase{  \begin{numcases}}
\def\ecase{\end{numcases}   \noindent}
\def\bsbcase{  \begin{subnumcases}}
\def\esbcase{\end{subnumcases}   \noindent}

\newtheorem{theorem}{Theorem}
\newtheorem{corollary}{Corollary}
\newtheorem{lemma}{Lemma}

\newtheorem{definition}{Definition}

\newenvironment{myproof}{\noindent{\em Proof:} \hspace*{1em}}{
    \hspace*{\fill} $\Box$ }
\newenvironment{proof_of}[1]{\noindent {\em Proof of #1: }}{\hspace*{\fill} $\Box$ }

\newcommand{\matplottc}[1]{               
        \unitlength .45truein
        \begin{center}
        \includegraphics{#1.ps}
        \end{picture}
        \end{center}
}

\def\psfancypar#1#2{\begingroup\def\par{\endgraf\endgroup\lineskiplimit=0pt}
               \setbox2=\hbox{\large\sc #2}
               \newdimen\tmpht \tmpht \ht2 \advance\tmpht by \baselineskip
               \font\hhuge=Times-Bold at \tmpht
               \setbox1=\hbox{{\hhuge #1}}
               \count7=\tmpht \count8=\ht1
               \divide\count8 by 1000 \divide\count7 by \count8
               \tmpht=.001\tmpht\multiply\tmpht by \count7
               \font\hhuge=Times-Bold at \tmpht
               \setbox1=\hbox{{\hhuge #1}}
               \noindent
                \hangindent1.05\wd1
               \hangafter=-2 {\hskip-\hangindent
               \lower1\ht1\hbox{\raise1.0\ht2\copy1}%
                \kern-0\wd1}\copy2\lineskiplimit=-1000pt}

\def\Kout{\setbox1=\hbox{\Huge\bf K}\hbox to
1.05\wd1{\hspace{.05\wd1}
}}
\def\Sout{\setbox1=\hbox{\Huge\bf S}\hbox to 1.05\wd1{\hspace{.05\wd1}
}}

\allowdisplaybreaks[4]

\newcommand{\bX}{\mathbf{X}}

\newcommand{\calN}{\mathcal{N}}

\newcommand{\bx}{\mathbf{x}}

\newcommand{\tilJ}{\widetilde{J}}

\DeclareMathOperator{\sign}{sign}

\title{High-Dimensional  Covariance  Decomposition\\ into Sparse Markov and Independence Models}

\author{Majid Janzamin and Animashree Anandkumar\footnote{M. Janzamin and A. Anandkumar  are with the Center for Pervasive Communications and Computing, Electrical Engineering and Computer Science Dept., University of California, Irvine, USA 92697. Email: mjanzami@uci.edu, a.anandkumar@uci.edu}}

\begin{document}

\maketitle

\begin{abstract} \noindent Fitting high-dimensional data involves   a delicate tradeoff between faithful representation and the use of sparse models. Too often, sparsity assumptions on the fitted model are too restrictive to provide a faithful representation of the observed data. In this paper, we present a novel framework incorporating  sparsity in different domains.
We decompose the observed covariance matrix into a sparse Gaussian Markov model (with a sparse precision matrix)  and a sparse independence model (with a sparse covariance matrix). Our framework incorporates sparse covariance and sparse precision estimation as special cases and thus introduces a richer class of high-dimensional models. 
We  characterize   sufficient conditions for identifiability  of the two models, \viz  Markov and   independence models. We propose an efficient decomposition method based on a  modification of the popular $\ell_1$-penalized maximum-likelihood estimator ($\ell_1$-MLE). We establish that our estimator is consistent in both the domains, i.e., it successfully recovers the supports of both   Markov and   independence models, when the number of samples $n$ scales as $n = \Omega(d^2 \log p)$, where $p$ is the number of variables and $d$ is the maximum node degree in the Markov model.
Our experiments validate these results and also demonstrate that our models have better inference accuracy under simple algorithms such as loopy belief propagation.
\end{abstract}



\section{Introduction}

Covariance   estimation is a  classical problem in multi-variate statistics.  The idea that  second-order statistics capture   important and relevant relationships between a given set of variables is natural.  Finding the sample covariance matrix based on observed data is straightforward and widely used~\cite{Anderson:book}. However, the sample covariance matrix  is ill-behaved in high-dimensions, where the number of dimensions $p$ is typically much larger than the number of available samples $n$ $(p\gg n)$. Here, the problem of covariance estimation is ill-posed since the  number of unknown parameters is larger than the number of available samples, and   the sample covariance matrix becomes singular in this regime.

Various solutions have been proposed for high-dimensional covariance estimation. Intuitively,   by restricting the class of covariance models to those with a limited number of free parameters, we can successfully estimate the models  in high dimensions. A natural mechanism to achieve this is to impose a sparsity constraint on the covariance matrix. In other words, it is presumed that there are only a few (off-diagonal) non-zero entries in the covariance matrix, which implies that the variables under consideration approximately satisfy {\em marginal independence}, corresponding to the zero pattern of the covariance matrix~\cite{kauermann1996dualization} (and we refer to such models as independence models). Many works have studied this setting and have provided guarantees for high-dimensional estimation through simple  thresholding of the sample covariance matrix and other related schemes. See Section~\ref{sec:related}. In many settings, however, marginal independence is too restrictive and does not hold. For instance, consider the dependence between the monthly stock returns of various companies listed on the S\&P 100 index. It is quite possible that a wide range of complex (and unobserved) factors such as the economic climate, interest rates etc., affect the returns of all the companies. Thus, it is not realistic to model the stock returns of various companies through a sparse covariance model.

A popular alternative sparse   model, based on {\em conditional independence} relationships, has gained widespread acceptance in recent years~\cite{Lauritzen:book}. In this case, sparsity is imposed {\em not} on the covariance matrix, but on the inverse covariance or the {\em precision} matrix. It can be shown that the zero pattern of the precision matrix corresponds to a set of conditional-independence relationships and such models are referred to as graphical or Markov models.  Going back to the stock market example, a first-order approximation is to model   the companies in different divisions\footnote{\scriptsize See \url{http://www.osha.gov/pls/imis/sic_manual.html} for classifications of the companies.} as conditionally independent given the S\&P 100 index variable, which captures the overall trends of the stock returns, and thus removes much of the dependence between the companies in different divisions. 
High-dimensional estimation in models with sparse precision matrices   has been widely studied, and guarantees for estimation have been provided under a set of sufficient conditions. See Section~\ref{sec:related} for related works.
However, sparse Markov models  may not be always sufficient to capture all the statistical relationships among variables. Going back to the stock market example, the approximation of using the S\&P   index node to capture the dependence between companies of different divisions may not be enough. For instance, there can still be a large {\em residual} dependence between the companies in manufacturing and mining divisions, which cannot be accounted by the S\&P index node. 

\begin{figure}\centering{
\bp\psfrag{Si}[l]{$\Sigma^*$}\psfrag{J}[l]{${J_M^*}^{-1}$}
\psfrag{SR}[l]{$\Sigma_R^*$}\psfrag{Sr}[l]{\tcr{$S_R$}}
\psfrag{SM}[l]{\tcdkb{$S_M$}}
\includegraphics[width=4in]{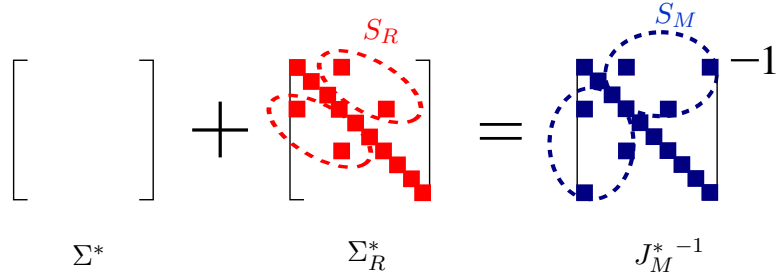}\ep}
\caption{\small Representation of the covariance decomposition problem, where perturbing the observed covariance matrix with a structured noise model results in a sparse graphical model. The case where the noise model has sparse marginal dependencies is considered.}
\label{fig:matrix}
\end{figure}

In this paper, we consider decomposition of the observed data into two domains, \viz Markov and independence domains. We posit that the observed data results in a sparse graphical model under structured perturbations in the form of an independence model, see Fig.\ref{fig:matrix}. This framework encapsulates Markov and independence models, and incorporates a richer class of models which can faithfully capture complex relationships, such as in the  stock market example above, and yet retain parsimonious representation.
The idea that a combination of Markov and independence models can provide  good model-fitting  is not by itself new and  perhaps the work which is closest to ours is~\cite{Choi&etal:10SP}, where   multi-resolution models with a known hierarchy of variables is considered. Their model consists of a combination of a  sparse precision matrix, which captures the conditional independence across scales, and  a sparse covariance matrix, which captures the residual in-scale correlations. Heuristics for learning and inference are provided in~\cite{Choi&etal:10SP}.
However, the approach in~\cite{Choi&etal:10SP} has several deficiencies, including lack of theoretical guarantees, assumption of a known sparsity support for the Markov model, use of expectation maximization (EM) which has no guarantees of reaching the global optimum, non-identifiability due to the presence of both latent variables and residual correlations, and so on. In contrast, we develop efficient convex optimization methods for decomposition, which are easily implementable and also provide theoretical guarantees for successful recovery. In summary,  in this paper, we provide an in-depth study of efficient methods and guarantees for joint estimation of a combination of Markov and independence models.

Our model reduces to sparse covariance and sparse inverse covariance estimation for certain choices of tuning parameter.  Therefore, we incorporate a range of models from sparse covariance to sparse inverse covariance.


%

\subsection*{Summary of Contributions}

We consider joint estimation of  Markov and  independence models, given observed data in a high dimensional setting. Our contributions in this paper are three fold. First, we derive a set of sufficient restrictions, under which there is a unique decomposition into the two domains, \viz the Markov and the independence domains, thereby leading to an {\em identifiable} model. Second, we propose   novel and   efficient estimators for obtaining the decomposition, under both exact and sample statistics. Third, we provide strong theoretical guarantees for high-dimensional learning, both in terms of norm guarantees and {\em sparsistency} in each domain, \viz the Markov and the independence domain. 

Our learning method is based on convex optimization.  We adapt the popular $\ell_1$-penalized maximum likelihood estimator (MLE),  proposed   originally for sparse Markov model selection and has efficient implementation in the form of graphical lasso \cite{Friedman&etal:07}.  This method involves an $\ell_1$ penalty   on the precision matrix, which is a convex relaxation of the $\ell_0$ penalty, in order to encourage sparsity in   the precision matrix. The Lagrangian dual of this program is a {\em maximum entropy} solution which approximately fits   the given sample covariance matrix. We modify this program to our setting as follows: we incorporate an additional $\ell_1$ penalty term involving the residual covariance matrix (corresponding to the independence model) in the max-entropy program. This term can be viewed as encouraging sparsity in the independence domain, while fitting a maximum entropy Markov model to the rest of the sample  correlations.
We characterize the optimal solution of the above program, and also provide intuitions on the class of Markov and independence model combinations  which can be incorporated under this framework. As a byproduct of this analysis, we obtain a set of conditions for identifiability of the two model components.

We provide strong theoretical guarantees for our proposed method under a set of sufficient conditions. We establish that it is possible to obtain {\em sparsistency} and norm guarantees in both the Markov and the independence domains. We establish that the number of samples $n$ is required to scale as $ n= \Omega(d^2 \log p)$ for consistency, where $p$ is the number of variables, and $d$ is the maximum degree in the Markov graph.
The set of sufficient conditions for successful recovery are based on the so-called notion of {\em mutual incoherence}, which controls the dependence between different sets of variables~\cite{Ravikumar&etal:08Arxiv}. In Section~\ref{sec:experiments}, the synthetic experiments are run on a model which does not necessarily satisfy sufficient mutual incoherence conditions; But we observe that our method has good numerical estimation performance even when the above incoherence conditions are not fully satisfied.

We establish that our estimation reduces to sparse covariance and sparse inverse covariance estimation for certain choices of tuning parameter. On one end, it reduces to the $\ell_1$ penalized MLE for sparse precision estimation~\cite{Ravikumar&etal:08Arxiv}. On the other extreme, it reduces to (soft) threshold estimator for sparse covariance estimator, on lines of~\cite{Bickel&Levina:08Stat}. Moreover, our conditions for successful recovery  are similar to those previously characterized for consistent estimation of sparse covariance/precision matrix.

Our experiments validate our theoretical results on the sample complexity and demonstrate  that our method is able to learn a richer class of models, compared to sparse graphical model selection, while requiring similar number of samples. In particular, our method is able to provide better estimates for the overall precision matrix, which is  dense in general, while
the performance of $\ell_1$-based optimization is worse since it attempts to approximate the dense matrix via a sparse estimate. Additionally, we demonstrate that our estimated models have better accuracy under simple distributed inference algorithms such as loopy
belief propagation (LBP). This is because the Markov components of the estimated models tend to be more {\em walk summable}~\cite{Malioutov&etal:06JMRL}, since some of the correlations can be ``transferred'' to the residual matrix. Thus, in addition to learning a richer model class, incorporating sparsity in both covariance and precision domains, we also learn models amenable to efficient inference. We also apply our method to real data sets. We see the resulting models are fairly interpretable for the real data sets. For instance, for stock returns data set, we observe in both Markov and residual graphs that there exist edges among companies in the same division or industry, e.g., in the residual graph, nodes ``HD", ``WMT", ``TGT" and ``MCD", all belonging to division Retail Trade form  a partition. Also for foreign exchange rate data set, we observe that the statistical dependencies of foreign exchange rates are correlated with the geographical locations of countries, e.g., it is observed in the learned model that the exchange rates of Asian countries are more correlated.

\subsection{Related Works}\label{sec:related}

There have been numerous works on high-dimensional  covariance selection and estimation, and we describe them below.
In all the settings below based on sparsity of the covariance matrix in some basis, the notion of consistent estimation of the sparse support is known as {\em sparsistency}.

\paragraph{Sparse Graphical Models:}Estimation of covariance matrices by exploiting the sparsity pattern in the inverse covariance or the precision matrix has a long history. The sparsity pattern of the precision matrix corresponds to a Markov graph of a graphical model which characterizes the set of conditional independence relationships between the variables.
Chow and Liu established that the maximum likelihood estimate (MLE) for tree graphical  models reduces to a maximum weighted spanning tree algorithm where the edge weights correspond to empirical mutual information.
The seminal work by Dempster~\cite{Dempster1972covariance} on covariance selection over chordal graphs analyzed the convex program corresponding to  the Gaussian MLE and its dual, when the graph structure is known.

In the high-dimensional regime, penalized likelihood methods have been used in a number of works  to achieve parsimony in covariance selection.   Penalized MLE based on $\ell_1$ penalty  has  been used in~\cite{huang2006covariance,Mei06,ABE:08,BEA:08,rothman2008sparse,Ravikumar&etal:08Arxiv}, among numerous other works,  where sparsistency and norm guarantees for recovery in high dimensions are provided. Graphical lasso \cite{Friedman&etal:07} is an efficient and popular implementation for the $\ell_1$-MLE. There have also been recent extensions to group sparsity structures\cite{yuan2006model,zhao2009composite}, scenarios  with missing samples~\cite{loh2011high} 
, semi-parametric settings based on non-paranormals~\cite{liu2009nonparanormal}, and to the non-parametric setting~\cite{kolar2010sparse}. In addition to the convex methods, there have also been a number of non-convex methods for Gaussian graphical model selection~\cite{spirtes1995learning,kalisch2007estimating,zhang2009consistency,AnandkumarTanWillsky:Gaussian11,Zhang:08NIPS}.
While we base much of our consistency analysis on~\cite{Ravikumar&etal:08Arxiv}, we also need to develop novel techniques to handle the delicate issue of errors in the two domains, \viz Markov and independence domains.

\paragraph{Sparse Covariance Matrices:}In contrast to the above formulation, alternatively we can  impose sparsity on the covariance matrix. Note that the zero pattern in the covariance matrix corresponds to   marginal independence relationships~\cite{cox1993linear,kauermann1996dualization,banerjee2003dualization}.
High-dimensional estimation of sparse covariance models has been extensively studied in ~\cite{el2008operator,Bickel&Levina:08Stat,cai2010optimal}, among others.
Wagaman and Levina~\cite{wagaman2009discovering} consider block-diagonal and banded covariance matrices and propose an Isomap method for discovering meaningful orderings of
variables. The work in~\cite{lam2009sparsistency} provides unified results for sparsistency under different sparsity assumptions, \viz sparsity in precision matrices, covariance matrices and models with sparse Cholesky decomposition.
\\
\\
The above works provide strong guarantees for covariance selection and estimation under various sparsity assumptions. However, they cannot handle matrices which are combinations of different sparse representations, but are otherwise dense when restricted to any single representation.


\paragraph{Decomposable Regularizers:}Recent works have considered model decomposition based on observed samples into desired parts through convex relaxation approaches.  Typically, each part is represented as an {\em algebraic variety}, which are based on {\em semi-algebraic} sets, and conditions for recovery of each component are characterized. For instance, decomposition of the inverse covariance matrix into sparse and low-rank varieties is considered in~\cite{Chandrasekaran&etal:10Siam,Chandrasekaran:10latent,Candes&etal:PCA} and is relevant for latent Gaussian graphical model.
The work in ~\cite{silva2011sparse} considers finding a sparse-approximation using a small number of positive semi-definite (PSD) matrices, where the ``basis'' or the set of PSD matrices is specified a priori.  In~\cite{negahban2010unified}, a unified framework is  provided for high-dimensional analysis of the so-called $M$-estimators, which optimize the sum of a convex  loss function with decomposable regularizers. A general framework for decomposition into a specified set of algebraic varieties was studied in~\cite{chandrasekaran2010convex}.
\\
\\
The above formulations, however, cannot incorporate our scenario, which consists of a combination of sparse Markov and independence graphs. This is because,
although  the constraints on the inverse covariance matrix (Markov graph) and the covariance matrix (independence graph) can each be specified in a straightforward manner,   their combined constraints on the resulting covariance matrix  is not easy to incorporate into a learning method. In particular, we do not have a decomposable regularizer for this setting.

\paragraph{Multi-Resolution Models:} Perhaps the work which is closest to ours is~\cite{Choi&etal:10SP}, where   multi-resolution models with a known hierarchy of variables is considered. The model consists of a combination of a  sparse precision matrix, which captures the conditional independence across scales, and  a sparse covariance matrix, which captures the residual in-scale correlations. Heuristics for learning and inference are provided. However, the work has three main deficiencies: the sparsity support  is assumed to be known, the proposed heuristics have no theoretical guarantees for success and the models considered are in general not identifiable, due to the presence of both latent variables and residual correlations.



\section{Preliminaries and Problem Statement}\label{sec:prelim}

\paragraph{Notation: } For any vector $v \in \mathbb{R}^p$ and  a real number $a \in [1,\infty)$, the notation $\| v \|_a$   refers to the $\ell_a$
norm of vector $v$ given by $\| v \|_a := \bigl( \sum_{i=1}^p |v_i|^a
\bigr)^\frac{1}{a}$. For any matrix $U \in \mathbb{R}^{p \times p}$, the
induced or the operator norm is given by $\gennorm{U}_{a,b} := \max_{\| z
\|_a = 1} \| Uz \|_b$ for parameters $a,b \in [1,\infty)$. Specifically,
we use the $\ell_\infty$ operator norm which is equivalent to
$\gennorm{U}_\infty = \max_{i=1,...,p} \sum_{j=1}^p |U_{ij}|$. We also
have $\gennorm{U}_1 = \gennorm{U^T}_\infty$. Another induced norm is the spectral norm $\gennorm{U}_2$ (or $\gennorm{U}$) which is equivalent to the maximum singular value of $U$. We also use the $\ell_\infty$ element-wise norm notation $\|U\|_\infty$ to refer to the maximum absolute value of the entries of $U$. Note that it is not a matrix norm but a norm on the vectorized form of the matrix.
The trace inner product of two matrices is denoted by $\langle U,V \rangle
:= \text{Tr}(U^T V) = \sum_{i,j} U_{ij} V_{ij}$. Finally, we use the usual
notation for asymptotics: $f(n) = \Omega(g(n))$ if $f(n) \geq c g(n)$ for
some constant $c>0$ and $f(n) = O(g(n))$ if $f(n) \leq c' g(n)$ for some
constant $c' < \infty$.

\subsection{Gaussian Graphical Models}

A Gaussian graphical model is a family of jointly Gaussian distributions which factor in accordance to a given  graph.
Given a graph $G=(V,E)$, with $V  = \{1,\ldots, p\}$, consider a vector of Gaussian random   variables $\bfX=[X_1, X_2, \ldots, X_p]$, where each node $i \in V$    is associated with a scalar Gaussian random variable $X_i$. A Gaussian graphical model Markov on $G$ has a   probability density function (pdf) that may be  parameterized as
\beq\label{eqn:gauss}f_{\bfX}(\bfx) \propto \exp\left[-\frac{1}{2} \bfx^T J \bfx + \bfh^T \bfx\right],\eeq  where $J$ is a positive-definite symmetric matrix   whose sparsity pattern corresponds to that of the graph $G$. More precisely,   \beq \label{eqn:sparsity} J(i,j)=0 \iff (i,j) \notin G.\eeq The matrix $J$ is known as  the   potential or concentration matrix, the non-zero entries $J(i,j)$ as the edge potentials, and the vector $\bfh$ as the potential vector.   The form of parameterization in \eqref{eqn:gauss} is known as the information form and   is related to the standard mean-covariance parameterization of the Gaussian distribution as
\[  \mubf = J^{-1} \bfh,\quad \Sigma=J^{-1},\] where $\mubf:=\Ebb[\bfX]$ is the mean vector and  $\Sigma:=\Ebb[(\bfX-\mubf)(\bfX-\mubf)^T]$ is the covariance matrix.

We say that a jointly Gaussian random vector  $\bX$  with joint pdf $ f(\bx)$ satisfies local Markov property with respect to a graph $G$ if
\begin{equation}
f(x_i|\bx_{\calN(i)}) = f(x_i|\bx_{V\setminus i})
\end{equation}
holds for all nodes $i \in V$, where $\calN(i)$ denotes the set of neighbors of node $i\in V$ and,  $V\setminus i$ denotes the set of all nodes   excluding $i$. More generally, we say that  $\bfX$  satisfies the global Markov property, if for all disjoint sets $A,B\subset V$, we have
\beq f(\bx_A, \bx_B|\bx_S) = f(\bx_A|\bx_S) f(\bx_B|\bx_S).\eeq where    set $S$ is a {\em separator}\footnote{A set $S\subset V$ is a separator for sets $A$ and $B$ if the removal of nodes in $S$ partitions  $A$ and $B$ into distinct components.} of $A$ and $B$. The local and global Markov properties are equivalent for non-degenerate Gaussian distributions~\cite{Lauritzen:book}.

On lines of the above description of graphical models, consider the class of Gaussian models\footnote{In the sequel, we denote the Markov graph, corresponding the support of the information matrix, as $G$ and the conjugate graph, corresponding to the support of the covariance matrix, as $G_c$.} $\Nc(\mu, \Sigma_{G_c})$, where the covariance matrix is supported on a graph $G_c$ (henceforth referred to as the conjugate graph), i.e., \beq \Sigma_{G_c}(i,j) = 0 \equiv (i,j) \notin G_c.\eeq Recall that uncorrelated Gaussian variables are independent, and thus, \beq X_i \indep X_j \equiv (i,j) \notin G_c.\eeq Equivalence between pairwise independence and global Markov properties were studied in~\cite{cox1993linear,kauermann1996dualization,banerjee2003dualization}.

In this paper, we posit that the observed model results in a sparse graphical model under structure perturbations in the form of an independence model: \beq\label{eqn:problemprelim} \Sigma^* + \Sigma_R^*= {J^*_M}^{-1}, \quad \Supp(J^*_M) = G_M, \Supp(\Sigma^*_R)=G_R,\eeq where $\Supp(\cdot)$ denotes the set of non-zero (off-diagonal) entries, $G_M$ denotes the Markov graph and $G_R$, the independence graph.

\subsection{Problem Statement}\label{sec:problem}
We now give a detailed description of our problem statement, which consists of the covariance decomposition problem (given exact statistics) and covariance estimation problem (given a set of samples).


\noindent{\bf Covariance Decomposition Problem: }
A fundamental question to be addressed is the identifiability of the model parameters.

\bd[Identifiability]A parametric model $\{P_\theta: \theta \in \Theta\}$ is identifiable with respect to a measure $\mu$ if there do not exist two distinct parameters $\theta_1\neq \theta_2$ such that $P_{\theta_1}=P_{\theta_2}$ almost everywhere with respect to $\mu$.\ed

Thus, if a model is not identifiable, there is no hope of estimating the model parameters from observed data.
A Gaussian graphical model (with no hidden variables) belongs to the family of standard exponential distributions~\cite[Ch. 3]{Wainwright&Jordan:08NOW}. Under non-degeneracy conditions, it is also in the minimal form, and as such is identifiable~\cite{Brown:book}.  In our setting in \eqref{eqn:problemprelim}, however, identifiability is not straightforward to address, and forms an important component of the    covariance decomposition problem, described below.

\noindent{\bf Decomposition Problem: }Given the covariance matrix $\Sigma^* = {J_M^*}^{-1} - \Sigma_R^*$ as in \eqref{eqn:problemprelim}, where $J_M^*$ is an unknown concentration matrix  and $\Sigma_R^*$ is an unknown  residual covariance  matrix, how and under what   conditions can we uniquely recover $J_M^*$ and $\Sigma_R^*$ from $\Sigma^*$?

In other words, we want to address whether the matrices $J_M^*$ and $\Sigma_R^*$ are {\em identifiable}, given $\Sigma^*$, and if so, how can we design efficient methods to recover them. If we do not impose any additional restrictions, there exists an {\em equivalence class} of models which form solutions to the decomposition problem.  For instance, we can model   $\Sigma^*$ entirely through an independence model $(\Sigma^* = \Sigma^*_R)$, or through a Markov model $(\Sigma^*= {J^*_M}^{-1})$. However, in most scenarios, these extreme cases are not desirable, since they   result in dense models, while we are interested in sparse representations with a parsimonious use of edges in both the graphs, \viz the Markov and the independence graphs. In Section~\ref{sec:Decomp_Assumptions}, we provide a sufficient set of structural and parametric conditions to guarantee identifiability of the Markov and the independence components, and in Section~\ref{sec: algo. formulation}, we propose an 
optimization program to obtain them.



\noindent{\bf Covariance Estimation Problem: }
In the above decomposition problem, we assume that the exact covariance matrix $\Sigma^*$ is known. However, in practice, we only have access to samples, and we describe this setting below.

Denote  $\widehat{\Sigma}^n$ as the sample covariance matrix\footnote{\scriptsize Without loss of generality, we limit our analysis to zero-mean Gaussian models. The results can be easily generalized to models with non-zero means.} \begin{equation}
\widehat{\Sigma}^n := \frac{1}{n} \sum_{k=1}^n x_{(k)} x_{(k)}^T,\label{eqn:samplecov}
\end{equation}
where $x_{(k)}, k=1,...,n$ are $n$ i.i.d. observations of a zero mean Gaussian random vector $X \sim \Nc(0, \Sigma^*)$, where $X:=(X_1,...,X_p)$. Now the estimation problem is described below.

\noindent{\bf Estimation Problem: }Assume that there exists a unique decomposition $\Sigma^* = {J_M^*}^{-1} - \Sigma_R^*$ where $J_M^*$ is an unknown concentration matrix with bounded entries and $\Sigma_R^*$ is an unknown sparse residual covariance  matrix given a set of constraints. Given the sample covariance matrix $\widehat{\Sigma}^n$, our goal is to find estimates of   $J_M^*$ and $\Sigma_R^*$ with provable guarantees.



In the sequel, we relate the exact and the sample versions of the decomposition problem.
In Section~\ref{sec:Sample Version Analysis},  we propose a modified  optimization program to obtain efficient estimates of the Markov and independence components. Under a set of sufficient conditions, we provide  guarantees   in terms of {\em sparsistency}, {\em sign consistency}, and {\em norm} guarantees, defined below.

\bd[Estimation Guarantees]We say that an estimate $(\hJ_M, \hSigma_R)$ to the decomposition problem in \eqref{eqn:problemprelim}, given a sample covariance matrix $\widehat{\Sigma}^n$, is    sparsistent or  model consistent, if the supports of $\hJ_M$ and $\hSigma_R$  coincide with the supports of ${J^*_M}$ and $\Sigma^*_R$ respectively.  It is said to be  sign consistent, if additionally, the respective signs coincide. The norm guarantees on the estimates is in terms of bounds on $\norm{\hJ_M - J_M^*}$ and $\norm{\hSigma_R - \Sigma^*_R}$, under some norm $\norm{\cdot}$. \ed

\section{Analysis  under Exact Statistics} \label{sec:exact statistics}

In this section, we provide the results under exact statistics.


\subsection{Conditions for Unique Decomposition} \label{sec:Decomp_Assumptions}


We first provide a set of sufficient conditions under which we can guarantee that the decomposition of $\Sigma^*$ in \eqref{eqn:problemprelim} into concentration matrix $J^*_M$ and residual matrix $\Sigma_R^*$ is unique\footnote{We drop the positive definite constraint on the residual matrix $\Sigma_R^*$ thereby allowing for a richer class of covariance decomposition. In section \ref{Structured Noise Model}, we modify the conditions and the learning method to incorporate positive definite  residual matrices $\Sigma_R^*$.}.
We impose the following set of constraints on the two matrices: \bi

\item[(A.0)] $\Sigma^*$ and $J_M^*$ are positive definite matrices, i.e., $\Sigma^* \succ 0, J_M^* \succ 0$.


\item[(A.1)] Off-diagonal entries of $J_M^*$ are  bounded from above, i.e., $\|J_M^*\|_{\infty, \operatorname{off}} \leq \lambda^*$, for some $\lambda^*>0$.
\item[(A.2)] Diagonal entries of $\Sigma_R^*$ are zero: $\bigl( \Sigma_R^* \bigr)_{ii} = 0$, and the support of its off-diagonal entries satisfies\beq \bigl( \Sigma_R^* \bigr)_{ij}\neq 0 \,\iff\, |\bigl( J_M^* \bigr)_{ij}| = \lambda^*, \quad \forall\,i\neq j. \eeq

\item[(A.3)] For any $i,j$, we have $\sign \bigl( \bigl( \Sigma_R^* \bigr)_{ij} \bigr) . \sign \bigl( \bigl( J_M^* \bigr)_{ij} \bigr) \geq 0$, i.e, the signs are the same.   \ei
 
Indeed, the above constraints restrict the class of models for which we can provide guarantees. However, in many scenarios, the above assumptions may be reasonable, and we now provide some justifications. (A.0) is a  natural assumption to impose since we are interested in valid $\Sigma^*$ and $J_M^*$ matrices. 
Condition (A.1) corresponds to bounded  off-diagonal entries of $J^*_M$. Intuitively, this limits the extent of ``dependence'' between the variables in the Markov model, and  can lead to models where inference can be performed with good accuracy using  simple algorithms such as belief propagation.
Condition (A.2) limits the support of the residual matrix $\Sigma^*_R$: the residual covariances are captured
at those locations (edges) where the concentration  entries $(J^*_M)_{i,j}$ are ``clipped" (i.e., the bound $\lambda^*$ is achieved). Intuitively, the Markov matrix $J^*_M$ is unable to capture all the correlations between the node pairs due to clipping, and the residual matrix $\Sigma^*_R$ captures the remaining correlations at the clipped locations.  Condition (A.3) additionally characterizes the signs of the  entries of $\Sigma^*_R$. For the special case, when the Markov model is attractive, i.e. $(J^*_M)_{i,j}\leq 0$  for $i \neq j$, the residual entries $(\Sigma^*_R)_{i,j}$ are also all negative. This implies that the model corresponding to $\Sigma^*$ is also attractive, since it only consists of positive correlations. By default, we set the diagonal entries of the residual matrix to zero in (A.2) and thus, assume that the Markov matrix captures all the variances in the model. 
In Section~\ref{sec:chainexample}, we provide a simple example of a Markov chain and a residual covariance model satisfying the above conditions. 

It is also worth mentioning that the number of model parameters satisfying above conditions is equivalent to the number of parameters in the special case of sparse inverse covariance estimation when $\lambda \rightarrow \infty$ \cite{Ravikumar&etal:08Arxiv}. It is assumed in assumption (A.2) that the residual matrix $\Sigma_R^*$ takes nonzero value when the corresponding entry in the Markov matrix $J_M^*$ takes its maximum absolute value $\lambda^*$. This assumption in conjunction with the sign assumption in (A.3), exactly determines the Markov entry $\bigl( J_M \bigr)_{ij}$ when the corresponding residual entry $\bigl( \Sigma_R \bigr)_{ij} \neq 0$. So, for each $(i,j)$ pair, only one of the entries $\bigl( J_M \bigr)_{ij}$ and $\bigl( \Sigma_R \bigr)_{ij}$ are unknown which results that the proposed model in this paper does not introduce additional parameters comparing to the sparse inverse covariance estimation, which is interesting.

According to the above discussion, we observe that the overall covariance and inverse covariance matrices $\Sigma^*$ and $J^*={\Sigma^*}^{-1}$ are dense, but represented with small number of parameters. It is interesting that we are able to represent models with dense patterns, but it is important to notice that the sparse representation leads to some restrictions on the model.

In the sequel, we propose an efficient method to recover the respective matrices $J^*_M$ and $\Sigma^*_R$ under conditions (A.0)-(A.3) and then establish the uniqueness of the decomposition. Finally, note that we do not impose any sparsity constraints on the concentration matrix $J^*_M$, and in fact, our method and guarantees allow for dense matrices $J^*_M$, when the exact covariance matrix $\Sigma^*$ is available. However, when only samples are available, we limit ourselves to sparse $J^*_M$ and provide learning guarantees in the high-dimensional regime, where the number of samples can be much smaller than the number of variables.

%

\subsection{Formulation of the Optimization Program} \label{sec: algo. formulation}

We now propose a 
method based on convex optimization for obtaining $(J^*_M, \Sigma^*_R)$ given the covariance matrix $\Sigma^*$ in \eqref{eqn:problemprelim}. Consider the following program
\begin{align}
\label{convex_prog_dual}
\bigl( \widehat{\Sigma}_M, \widehat{\Sigma}_R \bigr) &:= \argmax_{\Sigma_M \succ 0, \Sigma_R}  \log \det \Sigma_M - \lambda \|\Sigma_R\|_{1, \operatorname{off}} \\
& \operatorname{s.t.} \ \ \Sigma_M - \Sigma_R = \Sigma^*, \ (\Sigma_R)_d = 0 , \nonumber
\end{align}
where $\norm{\cdot}_{1, \off}$ denotes the $\ell_1$ norm of the off-diagonal entries, which is the sum of the absolute values of the off-diagonal entries, and $(\cdot)_d$ denotes the   diagonal entries.  Intuitively, the parameter $\lambda$ imposes a penalty on large residual covariances, and under favorable conditions, can encourage sparsity in the residual matrix. The program in \eqref{convex_prog_dual} can be recast\begin{align}
\label{convex_prog_dual_modified}&
\bigl( \widehat{\Sigma}_M, \widehat{\Sigma}_R \bigr) := \argmax_{\Sigma_M \succ 0, \Sigma_R}  \log \det \Sigma_M   \\
\operatorname{s.t.} \ \ & \ \Sigma_M - \Sigma_R = \Sigma^*, \ (\Sigma_R)_d = 0 , \|\Sigma_R\|_{1, \operatorname{off}}\leq C(\lambda),\nonumber
\end{align} for some constant $C(\lambda)$ depending on $\lambda$. The objective function in the above program corresponds to the entropy of the Markov model (modulo a scaling and a shift factor)~\cite{Cover&Thomas:book}, and thus, intuitively, the above program looks for the optimal Markov model with maximum entropy subject to an $\ell_1$ constraint on the residual matrix.

We declare the   optimal solution  $\hSigma_R$ in \eqref{convex_prog_dual}  as the estimate of the residual matrix $\Sigma_R^*$, and  $\hJ_M := \hSigma_M^{-1}$ as the estimate of the Markov concentration matrix $J^*_M$. The justification behind these estimates is based on the fact that the Lagrangian dual of the  program  in \eqref{convex_prog_dual} is (see Appendix~\ref{appendix:duality})
\begin{align}
\label{convex_prog_primal}
\widehat{J}_M := \argmin_{J_M \succ 0} & \langle \Sigma^*,J_M \rangle - \log \det J_M  \\
\operatorname{s.t.} & \ \|J_M\|_{\infty, \operatorname{off}} \leq \lambda , \nonumber
\end{align}
where $\norm{\cdot}_{\infty, \off}$ denotes the $\ell_{\infty}$ element-wise norm of the off-diagonal entries, which is the maximum absolute value of the off-diagonal entries.
Further, we show  in Appendix~\ref{appendix:duality}  that
the following relations exist between the optimal primal\footnote{Henceforth, we refer to the program in \eqref{convex_prog_primal} as the primal program and the program in \eqref{convex_prog_dual} as  the dual program.} solution $\widehat{J}_M$ and the optimal dual solution $\bigl( \widehat{\Sigma}_M,\widehat{\Sigma}_R \bigr)$: $\hJ_M = \hSigma_M^{-1}$, and thus, $\hJ_M^{-1} - \hSigma_R = \Sigma^*$ is a valid decomposition of the covariance matrix $\Sigma^*$.



\paragraph{Remark: }Notice that when the $\ell_\infty$ constraint is removed in the primal program in \eqref{convex_prog_primal}, which is equivalent to letting $\lambda\to \infty$,  the program corresponds to the maximum likelihood estimate, and the optimal solution in this case is $\hJ_M = {\Sigma^*}^{-1}$. Similarly, in the dual program in \eqref{convex_prog_dual}, when $\lambda \to \infty$,  the optimal solution corresponds to $\hSigma_M = \Sigma^* $ and $\hSigma_R=0$. At the other extreme, when $\lambda \to 0$, $\hJ_M$ is a diagonal matrix, and the residual matrix $\hSigma_R$ is in general, a full matrix (except for the diagonal entries). Thus, the parameter $\lambda$ allows us to carefully tune the contributions of the Markov and residual components, and we notice in our experiments in Section~\ref{sec:experiments} that $\lambda$ plays a crucial role in obtaining efficient decomposition into Markov and residual components.

\subsection{Guarantees and main results} \label{sec:Guarantees}


We now establish that the optimal solutions of the proposed   optimization programs in \eqref{convex_prog_dual}  and \eqref{convex_prog_primal} lead to a unique decomposition of the given covariance matrix $\Sigma^*$ under conditions (A.0)--(A.3) given in  Section~\ref{sec:Decomp_Assumptions}.

\begin{theorem}[Uniqueness of Decomposition]\label{thm:decomp} Under (A.0)--(A.3), given a covariance matrix $\Sigma^*$, if we set the parameter $\lambda = \|J_M^*\|_{\infty, \operatorname{off}}$ in the optimization program in \eqref{convex_prog_dual}, then the optimal solutions of primal-dual optimization programs \eqref{convex_prog_primal} and \eqref{convex_prog_dual} are given by $\bigl( \widehat{J}_M , \widehat{\Sigma}_R \bigr) = \bigl( J_M^* , \Sigma_R^* \bigr)$, and the decomposition is unique.
\end{theorem}

\bprf See Appendix~\ref{proof:decomp}.\eprf


Thus, we establish that the proposed optimization programs in \eqref{convex_prog_dual}  and \eqref{convex_prog_primal} {\em uniquely} recover the Markov concentration matrix $J_M^*$ and the residual covariance matrix $\Sigma_R^*$ given $\Sigma^*$ under conditions (A.0)--(A.3).

\section{Sample Analysis of the Algorithm} \label{sec:Sample Version Analysis}

In this section, we provide the results under sample statistics where some i.i.d. samples of random variables are only available.

\subsection{Optimization Program}

We have so far provided guarantees on unique decomposition given the exact covariance matrix $\Sigma^*$. We now consider the case, when $n$ i.i.d. samples are available from $\Nc(0, \Sigma^*)$, which allows us to estimate the sample covariance matrix $\widehat{\Sigma}^n$, as in \eqref{eqn:samplecov}.

We now modify the dual program in \eqref{convex_prog_dual}, considered in the previous section, to incorporate the sample covariance matrix $\widehat{\Sigma}^n$ as follows
\begin{align}
\label{convex_prog_dual_sample_case}
\bigl( \widehat{\Sigma}_M, \widehat{\Sigma}_R \bigr) := & \argmax_{\Sigma_M, \Sigma_R} \ \log \det \Sigma_M - \lambda \|\Sigma_R\|_{1, \operatorname{off}} \\
\operatorname{s.t.} & \ \|\widehat{\Sigma}^n - \Sigma_M + \Sigma_R\|_{\infty, \operatorname{off}} \leq \gamma  , \nonumber \\
& \ \bigl( \Sigma_M \bigr)_d = \bigl( \widehat{\Sigma}^n \bigr)_d, \ \bigl( \Sigma_R \bigr)_d = 0 \nonumber, \\
& \ \Sigma_M \succ 0, \Sigma_M - \Sigma_R \succ 0. \nonumber
\end{align}
Note that, in addition to substituting $\Sigma^*$ by $\widehat{\Sigma}^n$, there are two more modifications in the above program comparing to the exact case in \eqref{convex_prog_dual}.
First, the positive-definiteness constraint on the overall covariance matrix $\Sigma = \Sigma_M - \Sigma_R$ is added to make sure that the overall covariance matrix estimation is valid. This constraint is not required in the exact case since we have the constraint $\Sigma = \Sigma^*$ in that case which ensures the positive-definiteness of overall covariance matrix according to assumption (A.0) that $\Sigma^* \succ 0$.
Second, the equality constraint $\Sigma_M - \Sigma_R = \Sigma^*$ is relaxed on the off-diagonal entries by introducing the new parameter $\gamma$ which allows some deviation. More discussion including the Lagrangian primal form of the above optimization program and the effect of new parameter $\gamma$ is provided in section \ref{sec:proof outline}. 


\subsection{Assumptions under Sample Statistics} \label{sec:Assumption sample}

We now provide conditions under which we can provide guarantees for  estimating the Markov model $J_M^*$ and the residual model $\Sigma_R^*$, given the sample covariance $\widehat{\Sigma}^n$ in high dimensions. These are conditions in addition to conditions (A.0)--(A.3) in Section~\ref{sec:Decomp_Assumptions}.

The additional assumptions for successful recovery in high dimensions are based on the Hessian of the objective function in the optimization program in \eqref{convex_prog_primal_sample_case}, with respect to the variable $J_M$, evaluated at the true Markov model $J_M^*$. The Hessian of this function is given by~\cite{Boyd:book}
\begin{equation}\label{eqn:hessian}
\Gamma^* = {J_M^*}^{-1} \otimes {J_M^*}^{-1} = \Sigma_M^* \otimes \Sigma_M^*,
\end{equation}
where $\otimes$ denotes the Kronecker matrix product~\cite{Horn&Johnson:book}. Thus $\Gamma^*$ is a $p^2 \times p^2$ matrix indexed by the node pairs. Based on the results for exponential families~\cite{Brown:book}, $\Gamma^*_{(i,j),(k,l)} = \operatorname{Cov} \{X_i X_j, X_k X_l \}$, and hence it can be interpreted as an edge-based alternative to the usual covariance matrix $\Sigma_M^*$. Define $K_M$ as the $\ell_\infty$ operator norm  of the covariance matrix of the Markov model
\begin{equation}
K_M := \gennorm{\Sigma_M^*}_\infty.
\end{equation}
We now denote the supports of the Markov and residual models. Denote  $E_M := \{ (i,j) \in V \times V | i \neq j, \bigl( J_M^* \bigr)_{ij} \neq 0 \}$ as the edge set of Markov matrix $J_M^*$. Define
\begin{align}
& S_M := E_M \cup \{ (i,i) | i=1,...,p \} , \label{S_M definition} \\
& S_R := \{ (i,j) \in V \times V | \bigl( \Sigma_R^* \bigr)_{ij} \neq 0 \} . \label{Sigma_R Support def}
\end{align}
Thus, the set $S_M$ includes diagonal entries and also all edges of Markov graph corresponding to $J_M^*$. Also, recall from  (A.2) in Section~\ref{sec:Decomp_Assumptions} that the diagonal entries of $\Sigma^*_R$ are set to zero, and   that the support set $S_R$  is   contained in $S_M$, i.e.,  $S_R \subset S_M$. Let $S_M^c$ and $S_R^c$ denote the respective complement sets. Define \beq \label{set S def} S := S_M \cap S_R^c,\eeq so that $\{S_R, S, S_M^c\}$ forms a partition of   $\{ (1,...,p) \times (1,...,p) \}$. This partitioning plays a crucial role in being able to provide learning guarantees.
Define the maximum node degree for Markov model $J_M^*$ 
as
\begin{equation}
d:=\max_{j=1,...,p} |\{ i : (i,j) \in S_M \}|.
\end{equation}
Finally, for any two subsets $T$ and $T'$ of $V \times V$, $\Gamma^*_{TT'}$ denotes the submatrix of $\Gamma^*$ indexed by $T$ as rows and $T'$ as columns. We now impose various constraints on the submatrices of the Hessian  in \eqref{eqn:hessian}, limited to each of the sets  $\{S_R, S, S_M^c\}$.

\bi \item[(A.4)] \textbf{Mutual Incoherence}: These conditions impose mutual incoherence among three partitions of $\Gamma^*$ indexed by $S_R$, $S_M^c$ and $S$.
\begin{align}\label{eqn:incoherence1}
\max \{ \gennorm {\Gamma^*_{S_M^c S} \bigl( \Gamma^*_{S S} \bigr)^{-1} \Gamma^*_{S S_R} - \Gamma^*_{S_M^c S_R}}_{\infty} , \gennorm {\Gamma^*_{S_M^c S} \bigl( \Gamma^*_{S S} \bigr)^{-1}}_{\infty} \} & \leq (1 - \alpha) \ \operatorname{for} \ \operatorname{some} \ \alpha \in (0,1] , \\ \label{eqn:incoherence2}
K_{S S_R} := \gennorm {\bigl( \Gamma^*_{S S} \bigr)^{-1} \Gamma^*_{S S_R}}_{\infty} & < \frac{1}{4} .
\end{align}
\item[(A.5)] \textbf{Covariance Control}: For the same $\alpha$ specified above, we have the bound:
\begin{equation}\label{eqn:incoherence3}
K_{S S} := \gennorm {\bigl( \Gamma^*_{S S} \bigr)^{-1}}_{\infty} \leq \frac{(m-4) \alpha}{4(m - (m-1) \alpha)} \ \operatorname{for} \ \operatorname{some} \ m>4.
\end{equation}
\item[(A.6)] \textbf{Eigenvalue Control}: The minimum eigenvalue of overall covariance matrix $\Sigma^*$ satisfies the lower bound 
\beq \lambda_{\min} (\Sigma^*) \geq C_6 d \sqrt{ \frac{\log(4p^{\tau})}{n} } + C_7 d^2 \frac{\log(4p^{\tau})}{n} \ \operatorname{for} \ \operatorname{some} \ C_6,C_7 > 0 \ \operatorname{and} \ \tau>2.  \eeq
\ei

In (A.4), the condition in \eqref{eqn:incoherence1} bounds the effect of the non-edges of the Markov model, indexed by $S_M^c$, to its   edges,  indexed by $S_R$ and $S$. Note that we distinguish between the common edges of the Markov model with the residual model $(S_R)$ and the remaining edges of the Markov model $(S)$. The second condition in \eqref{eqn:incoherence2}   controls the influence of the edge-based terms which are shared  with the residual matrix, indexed by $S_R$,  to   other edges of the Markov model, indexed by $S$ = $S_M \cap S_R^c$.
Condition (A.5) imposes   $\ell_\infty$ bounds on the rows of $(\Gamma^*_{S S})^{-1}$. Note that for sufficiently large $m$, the bound in \eqref{eqn:incoherence3} tends to $\frac{\alpha}{4(1-\alpha)}$. Also note that the conditions (A.4) and (A.5) are only imposed on the Markov model $J^*_M$ and there are no additional constraints on the residual matrix $\Sigma^*_R$ (other than the conditions previously introduced in Section~\ref{sec:Decomp_Assumptions}). In condition (A.6), it is assumed that the minimum eigenvalue of overall covariance matrix $\Sigma^*$ is sufficiently far from zero to make sure that its estimation $\widehat{\Sigma}$ is positive definite and therefore a valid covariance matrix.

\subsubsection{Example of a Markov Chain + Residual  Covariance Model}\label{sec:chainexample}
\begin{figure}\centering{
\bp\psfrag{1}[l]{1}\psfrag{2}[l]{2}
\psfrag{3}[l]{3}\psfrag{4}[l]{4}
 \includegraphics[width=2in]{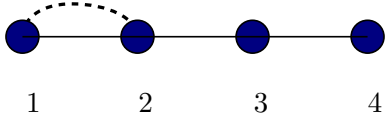}\ep}
\caption{\small Example of a Markov chain and a residual covariance matrix, where a residual edge is present between nodes 1 and 2.}\label{fig:markovchain}
\end{figure}

In this section, we propose a simple model satisfying assumptions (A.0)--(A.5). Consider a Markov chain with concentration matrix   $J_M^*$ over 4 nodes, as shown in Fig.\ref{fig:markovchain}. The diagonal entries in the corresponding covariance matrix $\Sigma_M^* =
{J_M^*}^{-1}$ are set to unity,  and the correlations between the neighbors in $J^*_M$ are set uniformly
to some value $\rho \in (-1,1)$, i.e., $\bigl( \Sigma_M^* \bigr)_{ij} =
\rho$ for $(i,j) \in E_M$. Due to the Markov property, the correlations between other node pairs are given by  $\bigl( \Sigma_M^* \bigr)_{13} = \bigl(
\Sigma_M^* \bigr)_{24} = \rho^2$ and $\bigl( \Sigma_M^* \bigr)_{14} =
\rho^3$. For the residual covariance matrix $\Sigma^*_R$, we consider one edge between nodes 1 and 2, i.e., $S_R = \{ (1,2),(2,1)\}$. It is easy to see that conditions (A.0)--(A.2) are satisfied.
Recall that $ S_M^c = \{(i,j):(i,j) \notin E_M \}$ and the remaining node pairs
belongs to set $S:= S_M \setminus S_R$. Through some straightforward calculations, we
can show that for any $| \rho | < 0.07$, the  mutual incoherence conditions in
(A.4) and (A.5) are satisfied for $\alpha=0.855$ and $m \geq 83$.
Note that the value of nonzero entries of $\Sigma_R^*$ are not
involved or restricted by these assumptions. However, they do need to satisfy the sign condition in (A.3). Thus, we have non-trivial models satisfying the set of sufficient conditions for successful high-dimensional estimation\,\footnote{Similarly, for the case when the correlations corresponding to Markov edges are distinct as $\bigl( \Sigma_M^* \bigr)_{12} =\rho_1, \bigl( \Sigma_M^* \bigr)_{23} =\rho_2$, and $\bigl( \Sigma_M^* \bigr)_{34} =\rho_3$, we can argue the same conditions. For compatibility with Figure~\ref{fig:markovchain}, assume that $\rho_1$ is the maximum among these three parameters, and therefore, the residual edge is between nodes 1 and 2. This is because the maximum of off-diagonal entries of $J_M^*$ also happens in entry $(1,2)$. Then, the same condition  $| \rho_1 | < 0.07$ is sufficient for satisfying conditions (A.0)--(A.5).}.
In Section~\ref{sec:experiments}, the synthetic experiments are run on a model which does not necessarily satisfy mutual incoherence conditions (A.4) and (A.5); But we observe that our method has good numerical estimation performance even when the above incoherence conditions are not fully satisfied.


\subsection{Guarantees and Main Results} \label{sec:Guarantees sample version}

We are now ready to provide the main result of this paper.

\noindent \begin{theorem}
\label{sample case theorem}
Consider a Gaussian distribution with covariance matrix  $\Sigma^*= {J^*_M}^{-1} - \Sigma_R^*$  satisfying conditions  (A.0)-(A.6). Given a sample covariance matrix $\widehat{\Sigma}^n$ using $n$ i.i.d. samples from the Gaussian model,  let $\bigl( \widehat{J}_M, \widehat{\Sigma}_R \bigr)$ denote the optimal solutions of the primal-dual pair  \eqref{convex_prog_primal_sample_case} and \eqref{convex_prog_dual_sample_case}, with parameters $\gamma = C_1 \sqrt{\log p / n}$ and $\lambda = \lambda^* + C_2 \sqrt{\log p / n}$ for some constants $C_1,C_2 > 0$, where $\lambda^* := \norm{J^*_M}_{\infty,\off}$. Suppose that $\bigl( \Sigma_R^* \bigr)_{\operatorname{min}} := \min_{(i,j) \in S_R} |\bigl( \Sigma_R^* \bigr)_{ij}|$ scales as $\bigl( \Sigma_R^* \bigr)_{\operatorname{min}} = \Omega \bigl( \sqrt{\log p / n} \bigr)$ and the sample size $n$ is lower bounded as
\begin{equation}
\label{asymptotic sample complexity}
n = \Omega \bigl( d^2 \log p \bigr),
\end{equation}
then with probability greater than $1-1/p^c \rightarrow 1$ (for some $c > 0$), we have:
\bi \item[a)] The estimates $\widehat{J}_M \succ 0$ and $\widehat{\Sigma}_R$ satisfy $\ell _{\infty}$ bounds
\begin{align}
& \|\widehat{J}_M - J_M^*\|_{\infty} = O \biggl( \sqrt{\frac{\log p}{n}} \biggr) , \\
& \|\widehat{\Sigma}_R - \Sigma_R^*\|_{\infty} = O \biggl( \sqrt{\frac{\log p}{n}} \biggr).
\end{align}

\item[b)]The estimate $\widehat{\Sigma}_R$ is sparsistent and sign consistent with $\Sigma_R^*$.

\item[c)] If in addition, $\bigl( J_M^* \bigr)_{\operatorname{min}} := \min_{(i,j) \in S_M} |\bigl( J_M^* \bigr)_{ij}|$ scales as $\bigl( J_M^* \bigr)_{\operatorname{min}} = \Omega \bigl(\sqrt{\log p / n} \bigr)$, then the estimate $\widehat{J}_M$ is   sparsistent and sign consistent with $J_M^*$.

\ei
\end{theorem}

\bprf See Appendix~\ref{appendix: proof of sample case theorem}.\eprf

\noindent{\bf Remarks: }\ben

\item {\bf Non-asymptotic sample complexity and error bounds: }In the above theorem, we establish that the number of samples is required to scale as $n = \Omega(d^2 \log p)$. In fact, our results are non-asymptotic, and the exact constants are provided in inequality \eqref{exact sample complexity}. The non-asymptotic form of error bounds are also provided in \eqref{J_M_error non-asymptotic} and \eqref{Sigma_R_error non-asymptotic}.

\item {\bf Extension to sub-Gaussian and other distributions: }In the above theorem, we considered Gaussian distribution. Similar to high dimensional covariance estimation in \cite{Ravikumar&etal:08Arxiv}, the result in the theorem can be easily extended to sub-Gaussian and other distributions with known tail conditions.

\item {\bf Comparison between direct estimation of $\Sigma^*$ and the above decomposition: }The overall matrix $\Sigma^*$ (and $J^*$) is a full matrix in general. Thus, if we want to estimate it directly, we need $n = \Omega \bigl( p^2 \log p \bigr)$ samples since the maximum node degree is $\Theta(p)$. Therefore, we can not estimate it directly in high dimensional regime and it demonstrates the importance of such sparse covariance + inverse covariance models for estimation.

\een

We discussed in the remark in section \ref{sec: algo. formulation} that the parameter $\lambda$ allows us to carefully tune the contributions of the Markov and residual components. When $\lambda \rightarrow \infty$, the program corresponds to $\ell_1$-penalized maximum likelihood estimator which is well-studied in \cite{Ravikumar&etal:08Arxiv,rothman2008sparse}. In this case, $\widehat{\Sigma}_R = 0$ and all the dependencies among random variables are captured by the sparse graphical model represented by $\widehat{J}_M$. On the other extreme, when $\lambda^*=0$ and thus $\lambda = C_2 \sqrt{\log p / n} \rightarrow 0$, with increasing the number of samples $n$, the off-diagonal entries in $\widehat{J}_M$ are bounded too tight by $\lambda$ (refer to the primal program in \eqref{convex_prog_primal_sample_case}) and therefore the residual covariance matrix $\widehat{\Sigma}_R$ captures most of the dependencies among random variables. In this case, we have the covariance estimation $\widehat{\Sigma} = \widehat{\Sigma}_M - \widehat{\Sigma}_R$, where the diagonal entries are included in $\widehat{\Sigma}_M$ and the off-diagonal entries are mostly included in $-\widehat{\Sigma}_R$.
In order to explain the results for these cases in a more concrete way, we explicitly mention the results for both sparse inverse covariance estimation ($\lambda \rightarrow \infty$) and sparse covariance estimation ($\lambda \approx 0$) methods in the following subsections. Note that both of these are special cases of the general result expressed in Theorem \ref{sample case theorem}. Thus, in Theorem \ref{sample case theorem}, we generalize these extreme cases to models with a linear combination of sparse covariance and sparse inverse covariance matrices.

\section{Discussions and Extension}

In this section, we first provide a detailed discussion of special cases sparse covariance and sparse inverse covariance estimation. Then, the extension of results to the structured noise model is mentioned.

\subsection{Sparse Inverse Covariance Estimation}

In this section, we mention the result for sparse inverse covariance estimation in high dimensional regime. This result is provided by \cite{Ravikumar&etal:08Arxiv} and is a special case of Theorem \ref{sample case theorem} when the parameter $\lambda$ goes to infinity. Before proposing the explicit result in Corollary \ref{Corollary: sparse precision est}, we state how the required conditions in Theorem \ref{sample case theorem} reduces to the conditions in \cite{Ravikumar&etal:08Arxiv}. \\ Since the support of residual matrix $\Sigma_R^*$ is a zero matrix in this special case, the mutual incoherence conditions in (A.4) reduce exactly to the same mutual incoherence condition in \cite{Ravikumar&etal:08Arxiv} as
\begin{equation}
\label{eqn:incoherence_Ravikumar}
\gennorm {\Gamma^*_{S^c S} \bigl( \Gamma^*_{S S} \bigr)^{-1}}_{\infty} \leq (1 - \alpha) \ \operatorname{for} \ \operatorname{some} \ \alpha \in (0,1] ,
\end{equation}
where $S=S_M$ is the support of Markov matrix $J^*=J_M^*$ as defined in \eqref{S_M definition}. Also note that the covariance control condition (A.5) is not required any more.\\
Furthermore, the sample complexity and convergence rate of $J_M^*$ estimation in Theorem \ref{sample case theorem} exactly reduce to the results in \cite{Ravikumar&etal:08Arxiv} as (for $q=8, l=3$)
\begin{align}
n > \overline{n}_f \Biggl( p^\tau ; 1 / \max \biggl \{ v_*, & 2 l d \Bigl( 1+\frac{q}{\alpha} \Bigr) K_{SS} K_M \max \Bigl \{ 1 , \frac{2}{l-1} \Bigl( 1+\frac{q}{\alpha} \Bigr) K_{SS} K_M^2 \Bigr\} \biggr \} \Biggr ), \label{sample complexity Ravikumar} \\
& \| \widehat{J} - J^* \|_{\infty} \leq  2 K_{S S} \Bigl( 1 + \frac{q}{\alpha} \Bigr) \overline{\delta}_f (p^\tau;n), \label{J_M_error Ravikumar}
\end{align}
where the result is valid for any $q \geq 8$ and $l>1$.

\noindent \begin{corollary}[Sparse Inverse Covariance Estimation, \cite{Ravikumar&etal:08Arxiv}] \label{Corollary: sparse precision est}
Consider a Gaussian distribution with covariance matrix  $\Sigma^*= {J^*}^{-1}$ satisfying mutual incoherence condition \eqref{eqn:incoherence_Ravikumar}. Given a sample covariance matrix $\widehat{\Sigma}^n$ using $n$ i.i.d. samples from the Gaussian model,  let $\widehat{J}$ denote the optimal solution of the primal-dual pair  \eqref{convex_prog_primal_sample_case} and \eqref{convex_prog_dual_sample_case}, with parameters $\gamma = C_1 \sqrt{\log p / n}$ and $\lambda \rightarrow \infty$ (removing $\ell_\infty$ constraints in the primal program \eqref{convex_prog_primal_sample_case}) for some constant $C_1 > 0$. Suppose that  the sample size $n$ is lower bounded as
\begin{equation}
n = \Omega \bigl( d^2 \log p \bigr),
\end{equation}
then with probability greater than $1-1/p^c \rightarrow 1$ (for some $c > 0$), we have:
\bi \item[a)] The estimate $\widehat{J} \succ 0$ satisfies $\ell _{\infty}$ bound
\begin{align}
& \|\widehat{J} - J^*\|_{\infty} = O \biggl( \sqrt{\frac{\log p}{n}} \biggr).
\end{align}
\item[b)]If in addition $\bigl( J^* \bigr)_{\operatorname{min}} := \min_{(i,j) \in S_M} |\bigl( J^* \bigr)_{ij}|$ scales as $\bigl( J^* \bigr)_{\operatorname{min}} = \Omega \bigl(\sqrt{\log p / n} \bigr)$, the estimate $\widehat{J}$ is sparsistent and sign consistent with $J^*$.
\ei
\end{corollary}

\textbf {Remark} [Comparison between the results of general case (Theorem \ref{sample case theorem}) and sparse inverse covariance estimation case (Corollary \ref{Corollary: sparse precision est})]: Considering the results in Theorem \ref{sample case theorem}, sample complexity and convergence rate of estimated models are exactly the same as results in \cite{Ravikumar&etal:08Arxiv} with only some minor
differences in coefficients. Compare \eqref{exact sample complexity} with \eqref{sample complexity Ravikumar} for sample complexity and \eqref{J_M_error non-asymptotic} with \eqref{J_M_error Ravikumar} for convergence rate of estimated Markov matrix $\widehat{J}_M$. But regarding the mutual incoherence conditions, we observe that the conditions for the special case sparse inverse covariance estimation in \eqref{eqn:incoherence_Ravikumar} are less restrictive than the conditions for the general case in \eqref{eqn:incoherence1}-\eqref{eqn:incoherence2}. Since the sparse inverse covariance estimation \cite{Ravikumar&etal:08Arxiv} is a special case of the general model in this paper, this additional limitation on models is inevitable, i.e., it is natural that we need some more incoherence conditions in order to be able to recover both the Markov and residual models in the general case.


\subsection{Sparse Covariance Estimation}

High-dimensional estimation of sparse covariance models has been studied in \cite{Bickel&Levina:08Stat}. They propose an estimation of a class of sparse covariance matrices by ``hard thresholding". They also prove spectral norm guarantees on the error between the estimated and exact covariance matrices. We also recover similar results in the other extreme case of proposed program \eqref{convex_prog_dual_sample_case} when $\lambda \approx 0$. The program reduces to the sparse covariance estimator as discussed earlier. In order to see that again, let us investigate the dual program restated as follows
\begin{align}
\bigl( \widehat{\Sigma}_M, \widehat{\Sigma}_R \bigr) := & \argmax_{\Sigma_M, \Sigma_R} \ \log \det \Sigma_M - \lambda \|\Sigma_R\|_{1, \operatorname{off}} \nonumber \\
\operatorname{s.t.} & \ \|\widehat{\Sigma}^n - \Sigma_M + \Sigma_R\|_{\infty, \operatorname{off}} \leq \gamma  , \nonumber \\
& \ \bigl( \Sigma_M \bigr)_d = \bigl( \widehat{\Sigma}^n \bigr)_d, \ \bigl( \Sigma_R \bigr)_d = 0 \nonumber, \\
& \ \Sigma_M \succ 0, \Sigma_M - \Sigma_R \succ 0. \nonumber
\end{align}
When the parameter $\lambda \approx 0$, the variable $\Sigma_R$ is very slightly penalized in the objective function. Therefore, most of the statistical dependencies are captured by $\Sigma_R$ and thus, off-diagonal entries of $\Sigma_M$ take very small values. Furthermore, according to the property of optimization program that the support of $\Sigma_R$ is contained within the support of $J_M$, sparsity on $\Sigma_R$ is encouraged by the effect of parameter $\gamma$. \\ It is also observed that we are approximately performing ``soft thresholding" in program \eqref{convex_prog_dual_sample_case} (when $\lambda \approx 0$) comparing to ``hard thresholding" in \cite{Bickel&Levina:08Stat}. Consider the case $\lambda=0$, where the Markov part $\Sigma_M$ is a diagonal matrix. Therefore, the $\|\widehat{\Sigma}^n - \Sigma_M + \Sigma_R\|_{\infty, \operatorname{off}} \leq \gamma$ constraint in the dual program \eqref{convex_prog_dual_sample_case} reduces to $\|\widehat{\Sigma}^n + \Sigma_R\|_{\infty, \operatorname{off}} \leq \gamma$ where it is seen that the negative soft thresholding is performed on matrix $\widehat{\Sigma}^n$ with threshold parameter $\gamma$, given by
\begin{equation}
S_{\gamma}(x) = \operatorname{sign}(-x) (|x|-\gamma)_{+}.
\end{equation}
Notice that we need to have $\lambda \approx 0$ for recovering the sparse covariance matrix given empirical covariances and in this case, we can view the estimator as approximately  performing soft thresholding.

Finally, we propose the corollary for this special case. Before that, we need some additional definitions for a general covariance matrix $\Sigma^*$. Similar to definition \eqref{Sigma_R Support def}, the support of a covariance matrix $\Sigma^*$ is defined as
\beq
S_{\Sigma} := \{ (i,j) \in V \times V | \Sigma^*_{ij} \neq 0 \}.
\eeq
The maximum node degree for a covariance matrix $\Sigma^*$ is also defined as
\begin{equation}
d_\Sigma:=\max_{j=1,...,p} |\{ i : (i,j) \in S_\Sigma \}|.
\end{equation}

\begin{corollary}[Sparse Covariance Estimation] \label{Corollary: sparse cov est}
Consider a Gaussian distribution with covariance matrix  $\Sigma^*$ satisfying eigenvalue control condition (A.6). Given a sample covariance matrix $\widehat{\Sigma}^n$ using $n$ i.i.d. samples from the Gaussian model,  let $\bigl( \widehat{\Sigma}_M, \widehat{\Sigma}_R \bigr)$ denote the optimal solutions of the primal-dual pair  \eqref{convex_prog_primal_sample_case} and \eqref{convex_prog_dual_sample_case}, with parameters $\gamma = C_1 \sqrt{\log p / n}$ and $\lambda = C_2 \sqrt{\log p / n}$ for some constants $C_1,C_2 > 0$. The estimated covariance matrix $\widehat{\Sigma}$ is defined as $\widehat{\Sigma}_{\operatorname{off}} := -\widehat{\Sigma}_R$ and $\widehat{\Sigma}_{d} := \bigl( \widehat{\Sigma}_M \bigr)_{d}$. Suppose that $\bigl( \Sigma_{\operatorname{off}}^* \bigr)_{\operatorname{min}} := \min_{(i,j) \in S_{\Sigma}, i \neq j} |\bigl( \Sigma^* \bigr)_{ij}|$ scales as $\bigl( \Sigma_{\operatorname{off}}^* \bigr)_{\operatorname{min}} = \Omega \bigl( \sqrt{\log p / n} \bigr)$ and the sample size $n$ is lower bounded as
\begin{equation}
n = \Omega \bigl( d_\Sigma^2 \log p \bigr),
\end{equation}
then with probability greater than $1-1/p^c \rightarrow 1$ (for some $c > 0$), we have:
\bi \item[a)] The estimate $\widehat{\Sigma}$ satisfies $\ell _{\infty}$ bound
\begin{align}
\|\widehat{\Sigma} - \Sigma^*\|_{\infty,\operatorname{off}} = O \biggl( \sqrt{\frac{\log p}{n}} \biggr). 
\end{align} 
\item[b)]The estimate $\widehat{\Sigma}_{\operatorname{off}}$ is sparsistent and sign consistent with $\Sigma^*_{\operatorname{off}}$.
\ei
\end{corollary}

\bprf
See Appendix \ref{Appendix:proof of sparse cov corollary}.
\eprf

\subsection{Structured Noise Model} \label{Structured Noise Model}
In the discussion up to now, we considered general   residual matrices $\Sigma_R^*$, not necessarily positive definite, thereby allowing for a rich  class of covariance decomposition models. In this section, we modify the conditions and the learning method to incorporate positive-definite residual matrices $\Sigma_R^*$. 

We   regularize the diagonal entries in an appropriate way to ensure that both $J_M^*$ and $\Sigma_R^*$ are positive definite. Thus, the identifiability assumptions  (A.0)-(A.3) are modified as follows:
\bi
\item[(A.0')] $\Sigma^*$, $\Sigma_R^*$ and $J_M^*$ are positive definite matrices, i.e., $\Sigma^* \succ 0, \Sigma_R^* \succ 0, J_M^* \succ 0$.
\item[(A.1')] $J_M^*$ is normalized such that $\bigl( J_M^* \bigr)_d = \lambda_1^*$ for some $\lambda_1^*>0$ and off-diagonal entries of $J_M^*$ are  bounded from above, i.e., $\|J_M^*\|_{\infty, \operatorname{off}} \leq \lambda_2^*$, for some $\lambda_2^*>0$.
\item[(A.2')] The off-diagonal entries of $\Sigma_R^*$ satisfy \beq \bigl( \Sigma_R^* \bigr)_{ij}\neq 0 \,\iff\, |\bigl( J_M^* \bigr)_{ij}| = \lambda_2^*, \quad \forall\,i\neq j. \eeq
\item[(A.3')] For any $i,j$, we have $\sign \bigl( \bigl( \Sigma_R^* \bigr)_{ij} \bigr) . \sign \bigl( \bigl( J_M^* \bigr)_{ij} \bigr) \geq 0$, i.e, the signs are the same.   \ei
It is seen in (A.1') that we put additional restrictions on diagonal entries of the Markov matrix $J_M^*$ in order to have nonzero diagonal entries for the residual matrix $\Sigma_R^*$. \\
Similar to the general form of dual program introduced in \eqref{convex_prog_dual_generalized}, we propose the following optimization program to estimate the Markov and residual components in the structured noise model:
\begin{align}
\bigl( \widehat{\Sigma}_M, \widehat{\Sigma}_R \bigr) := \argmax_{\Sigma_M, \Sigma_R \succ 0} & \log \det \Sigma_M - \lambda_1 \| \Sigma_R \| _{1, \operatorname{on}} - \lambda_2 \| \Sigma_R \| _{1, \operatorname{off}} \nonumber \\
\operatorname{s.t.} \ \ & \ \|\widehat{\Sigma}^n + \Sigma_R - \Sigma_M\|_{\infty, \operatorname{off}} \leq \gamma   , \\
& \ \bigl( \widehat{\Sigma}^n \bigr)_d + \bigl( \Sigma_R \bigr)_d =  \bigl( \Sigma_M \bigr)_d . \nonumber
\end{align}
The decomposition result under exact statistics can be similarly proven by setting parameter $\gamma=0$ when the identifiability assumptions (A.0')-(A.3') are satisfied. Furthermore, under additional estimation assumptions (A.4)-(A.6), the sample statistics guarantees in Theorem \ref{sample case theorem} can be also extended to the solutions of above program.

\section{Proof Outline} \label{sec:proof outline}

In this section, the Lagrangian primal form for the proposed dual program \eqref{convex_prog_dual_sample_case} is provided first and then the proof outlne is presented. For now, we drop the positive-definiteness constraint $\Sigma_M - \Sigma_R \succ 0$ in the proposed dual program \eqref{convex_prog_dual_sample_case}. We finally show that this constraint is satisfied for the proposed estimation under specified conditions and thus this constraint can be dropped. In the subsequent discussion, we drop this constraint. It is shown in Appendix \ref{appendix:duality} that the primal form for this reduced dual program is
\begin{align}
\label{convex_prog_primal_sample_case}
\widehat{J}_M := \argmin_{J_M \succ 0} \ & \langle \widehat{\Sigma}^n,J_M \rangle - \log \det J_M + \gamma \|J_M\|_{1, \operatorname{off}}  \\
\operatorname{s.t.} & \ \|J_M\|_{\infty, \operatorname{off}} \leq \lambda , \nonumber
\end{align}
We further establish that $\widehat{\Sigma}_M = \widehat{J}_M^{-1}$ is valid between the dual variable $\Sigma_M$ and primal variable $J_M$ and thus, \beq \label{eqn:decompsamples}\norm{ \widehat{\Sigma}^n - \widehat{J}_M^{-1}+\widehat{\Sigma}_R }_{\infty, \off}\leq   \gamma. \eeq
Comparing the above with the exact decomposition $\Sigma^* = {J^*_M}^{-1} - \Sigma_R^*$ in \eqref{eqn:problemprelim},  we note that for the sample version, we do not exactly fit the Markov and the residual models with the sample covariance matrix $\widehat{\Sigma}^n$, but allow for some divergence, depending on $\gamma$. Similarly, the primal program \eqref{convex_prog_primal_sample_case}  has an additional $\ell_1$ penalty term on $\hJ_M$, which is absent in \eqref{convex_prog_primal}. Having a non-zero $\gamma$ in the primal program enables us to impose a  sparsity constraint on $\hJ_M$, which in turn, enables us to estimate the matrices  in the  high dimensional regime $(p\gg n)$, under a set of conditions of sufficient conditions given in section \ref{sec:Assumption sample}.

We now provide a  high-level description of the proof for Theorem~\ref{sample case theorem}. The detailed proof is given in Appendix~\ref{appendix: proof of sample case theorem}. The proof  is based on the primal-dual witness method, which has been previously employed in~\cite{Ravikumar&etal:08Arxiv} and other works. However, we require significant modifications of this approach in order to handle the more complex setting of covariance decomposition.

In the primal-dual witness method, we define a modified version of the original optimization program \eqref{convex_prog_primal_sample_case}. Note that the key idea in constructing the modified version is to be able to analyze it and prove guarantees for it in a less complicated way comparing to the original version. Let us denote the solutions of the modified program by $\bigl( \widetilde{J}_M , \widetilde{\Sigma}_R \bigr)$ pair. In general, the optimal solutions of the two programs, original and modified one, are different. However,  under   conditions (A.0)--(A.5), we establish that their optimal solutions coincide. See Appendix~\ref{appendix: proof of sample case theorem} for details. Through this equivalence,  we thus establish that  the optimal solution $\bigl( \widehat{J}_M , \widehat{\Sigma}_R \bigr)$ of the original program in \eqref{convex_prog_primal_sample_case}  inherits all the properties of the optimal solution $\bigl( \widetilde{J}_M , \widetilde{\Sigma}_R \bigr)$ of the modified program, i.e., the solutions of the modified program act as witness for the original program. In the following, we define the modified optimization program and its properties. The primal-dual witness method steps which guarantee the equivalence between solutions of the original and the modified program are mentioned in Appendix~\ref{appendix: proof of sample case theorem}.

\begin{figure}\centering{
\bp\psfrag{S}[l]{$S$}
\psfrag{SR}[l]{$S_R$}
\psfrag{SMc}[l]{$S_M^c$}
\includegraphics[width=1.5in]{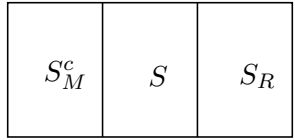}\ep}
\caption{\small The sets $S_R$, $S$ and $S_M^c$ form a partition of $\{ (1,...,p) \times (1,...,p) \}$, where $p$ is the number of nodes, $S_R$ is the support of the residual covariance matrix $\Sigma^*_R$ and $S_M$ is the support of the precision matrix $J^*_M$ of the Markov model and $S^c_M$ is its complement.}\label{fig:partition}
\end{figure}

We modify  the sample version of our optimization program in \eqref{convex_prog_primal_sample_case} as follows:
\begin{align}
\label{convex_prog_primal_sample_case_modified}
\widetilde{J}_M := \argmin_{J_M \succ 0} & \ \langle \widehat{\Sigma}^n,J_M \rangle - \log \det J_M + \gamma \|J_M\|_{1, \operatorname{off}}  \\
\operatorname{s.t.} & \ \bigl( J_M \bigr)_{S_M^c} = 0 , \ \bigl( J_M \bigr)_{S_R} = \lambda \sign \Bigl( \bigl( J_M^* \bigr)_{S_R} \Bigr) . \nonumber 
\end{align}Note that since we do not a priori know the supports of the original matrices $J_M^*$ and $\Sigma_R^*$, the above program cannot be implemented in practice, but is only  a device useful for proving consistency results. We observe  that the objective function in the modified program above  is the same as the original program in \eqref{convex_prog_primal_sample_case}, and only the constraints on the precision matrix are different in the two programs. In the above program in \eqref{convex_prog_primal_sample_case_modified}, constraints on the entries of the precision matrix when limited to sets $S_R$ and $S_M^c$ are more restrictive, while those in set $S:= S_M \setminus S_R$ are  more relaxed (i.e., the $\ell_\infty$ constraints present in \eqref{convex_prog_primal_sample_case} are removed above), compared to the original program in \eqref{convex_prog_primal_sample_case}.
Recall that $S_M$ denotes the support of the Markov model, while $S_R \subseteq S_M$ denotes the support of the residual or the independence model. See Fig.\ref{fig:partition}.

We now discuss the properties of the optimal solution $\bigl( \widetilde{J}_M , \widetilde{\Sigma}_R \bigr)$ of the modified program in \eqref{convex_prog_primal_sample_case_modified}.  Since the precision matrix entries on $S_M^c$ are set to zero in \eqref{convex_prog_primal_sample_case_modified}, we have that $\Supp(\tilJ_M) \subseteq \Supp(J^*_M)$. Denoting $\tilSigma_R$ as the residual covariance matrix corresponding to the modified program  \eqref{convex_prog_primal_sample_case_modified}, we can similarly characterize it in the following form derived from duality:
\begin{equation}
\bigl( \widetilde{\Sigma}_R \bigr)_{ij}= \left\{ \begin{array}{lcl}
		0 & \operatorname{for} & (i,j) \in S \\
		\widetilde{\beta}_{ij} & \operatorname{for} & (i,j) \in S_R,S_M^c ,
		\end{array}\right.
\label{SigmaR_JM_relation_sample_version_modified}
\end{equation}
where $\widetilde{\beta}_{ij}$ are the Lagrangian multipliers corresponding to the  equality constraints in the modified program \eqref{convex_prog_primal_sample_case_modified}.

%

Define   estimation errors $\widetilde{\Delta}_J := \widetilde{J}_M - J_M^*$ and $\widetilde{\Delta}_R := \widetilde{\Sigma}_R - \Sigma_R^*$ for the modified program in \eqref{convex_prog_primal_sample_case_modified}. It is easy to see that   $\bigl( \widetilde{\Delta}_J \bigr)_{S_R} = \lambda_\delta$, $\bigl( \widetilde{\Delta}_J \bigr)_{S_M^c} = 0$, $\bigl( \widetilde{\Delta}_R \bigr)_S = 0$, where $\lambda_\delta:=\lambda-\lambda^*>0$. This implies that   in any of the three sets $S$, $S_R$ or $S_M^c$, only one of the two  estimation errors $\widetilde{\Delta}_J$ or $\widetilde{\Delta}_R$ can be non-zero (or is at most $\lambda_\delta$). This property is crucial to be able to decouple the perturbations in the Markov and the independence domains, and thereby gives bounds on the individual perturbations. It is not clear if there is an alternative partitioning of the variables (here the partition is $S$, $S_R$ and $S_M^c$) which allows us to decouple the estimation errors for $\tilJ_M$ and $\tilSigma_R$. Through this decoupling, we are able to provide bounds on estimation errors $\widetilde{\Delta}_J$ and $\widetilde{\Delta}_R$ and thus, Theorem~\ref{sample case theorem} is established.


\section{Experiments}\label{sec:experiments}

In this section, we provide synthetic and real experimental results for the proposed algorithm.  We term our proposed optimization program 
as $\ell_1 + \ell_\infty$ method and compare it with the well-known $\ell_1$ method which is a special case of the proposed algorithm when $\lambda = \infty$. The primal optimization program \eqref{convex_prog_primal_sample_case} is implemented via the ADMM\,\footnote{Alternating Direction Method of Multipliers} technique proposed in~\cite{KarthikADMM:13Arxiv}.
We also compare the performance of belief propagation on the proposed model.

\subsection{Synthetic Data}
We build a Markov + residual synthetic model in the following way. We choose 0.2 fraction of Markov edges randomly to introduce residual edges. The underlying graph for the Markov part is a $q \times q$ 2-D grid structure (4-nearest neighbor grid). Therefore, the number of nodes is $p=q^2$. Because of assumption (A.2), we randomly set 0.2 fraction of nonzero Markov off-diagonal entries to $\{-0.2,0.2\}$, and the rest of nonzero off-diagonal  entries in $J_M^*$ (corresponding to the grid edges) are randomly chosen from set $\pm[0.15,0.2]$, i.e., $\bigl( J_M^* \bigr)_{ij} \in [-0.2,-0.15] \cup [0.15,0.2]$, for all $(i,j) \in E_M$. Note that 0.2 fraction of edges take the maximum absolute value which is needed by assumption (A.2).
Then we ensure that $J_M^*$ is positive definite by adding some uniform diagonal weighting. 
The nonzero entries of $\Sigma_R^*$ are chosen from $\pm[0.15,0.2]$ such that the sign of residual entry is the same as the sign of overlapping Markov entry (assumption (A.3)). We also generate a random mean in the interval $[0,1]$ for each variable. Note that this generated synthetic model does not necessarily satisfy mutual incoherence conditions (A.4) and (A.5); But we observe in the following that our method has good numerical estimation performance even when the incoherence conditions are not fully satisfied.

Before we provide experiment results, it is worth mentioning that the realization of above model is an example that both Markov and residual matrices $J_M^*$ and $\Sigma_R^*$ are sparse, while the overall covariance matrix $\Sigma^* = {J_M^*}^{-1} - \Sigma_R^*$ and concentration matrix $J^* = {\Sigma^*}^{-1}$ are both dense matrices.

\subsubsection*{Effect of graph size $p$} 
\begin{figure*}[t]\centering
\subfloat[a][]{\begin{minipage}{2.4in}
\bp
\psfrag{p eqq 25}[l]{\tiny $p\!=\!25$}
\psfrag{p eqq 64}[l]{\tiny $p\!=\!64$}
\psfrag{p eqq 100}[l]{\tiny $p\!=\!100$}
\psfrag{p eqq 400}[l]{\tiny $p\!=\!400$}
\psfrag{p eqq 900}[l]{\tiny $p\!=\!900$}
\psfrag{n}[l]{\scriptsize $n$}
\psfrag{Grid graph}[l]{\scriptsize Grid graph}
\psfrag{Distance between J}[l]{\scriptsize Norm. Dist\,$(\widehat{J}_M,J_M^*)$ }
\includegraphics[width=2.4in,height=1.3in]{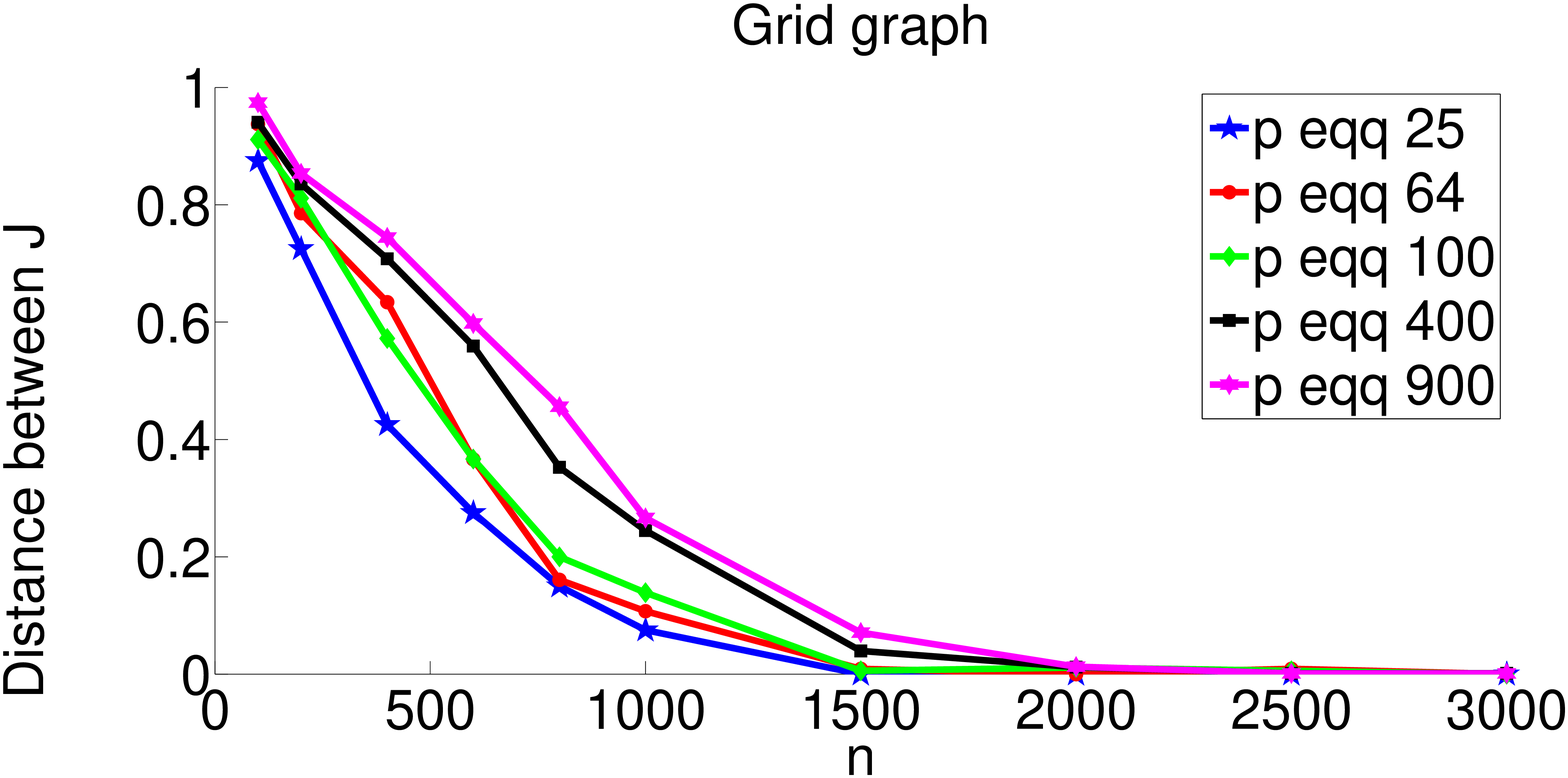}\ep
\end{minipage}}
\hfil
\subfloat[b][]{\begin{minipage}{2.4in}
\bp
\psfrag{p eqq 25}[l]{\tiny $p\!=\!25$}
\psfrag{p eqq 64}[l]{\tiny $p\!=\!64$}
\psfrag{p eqq 100}[l]{\tiny $p\!=\!100$}
\psfrag{p eqq 400}[l]{\tiny $p\!=\!400$}
\psfrag{p eqq 900}[l]{\tiny $p\!=\!900$}
\psfrag{n/log p}[l]{\scriptsize $n/\log p$}
\psfrag{Grid graph}[l]{\scriptsize Grid graph}
\psfrag{Distance between J}[l]{\scriptsize Norm. Dist\,$(\widehat{J}_M,J_M^*)$ }
\includegraphics[width=2.4in,height=1.3in]{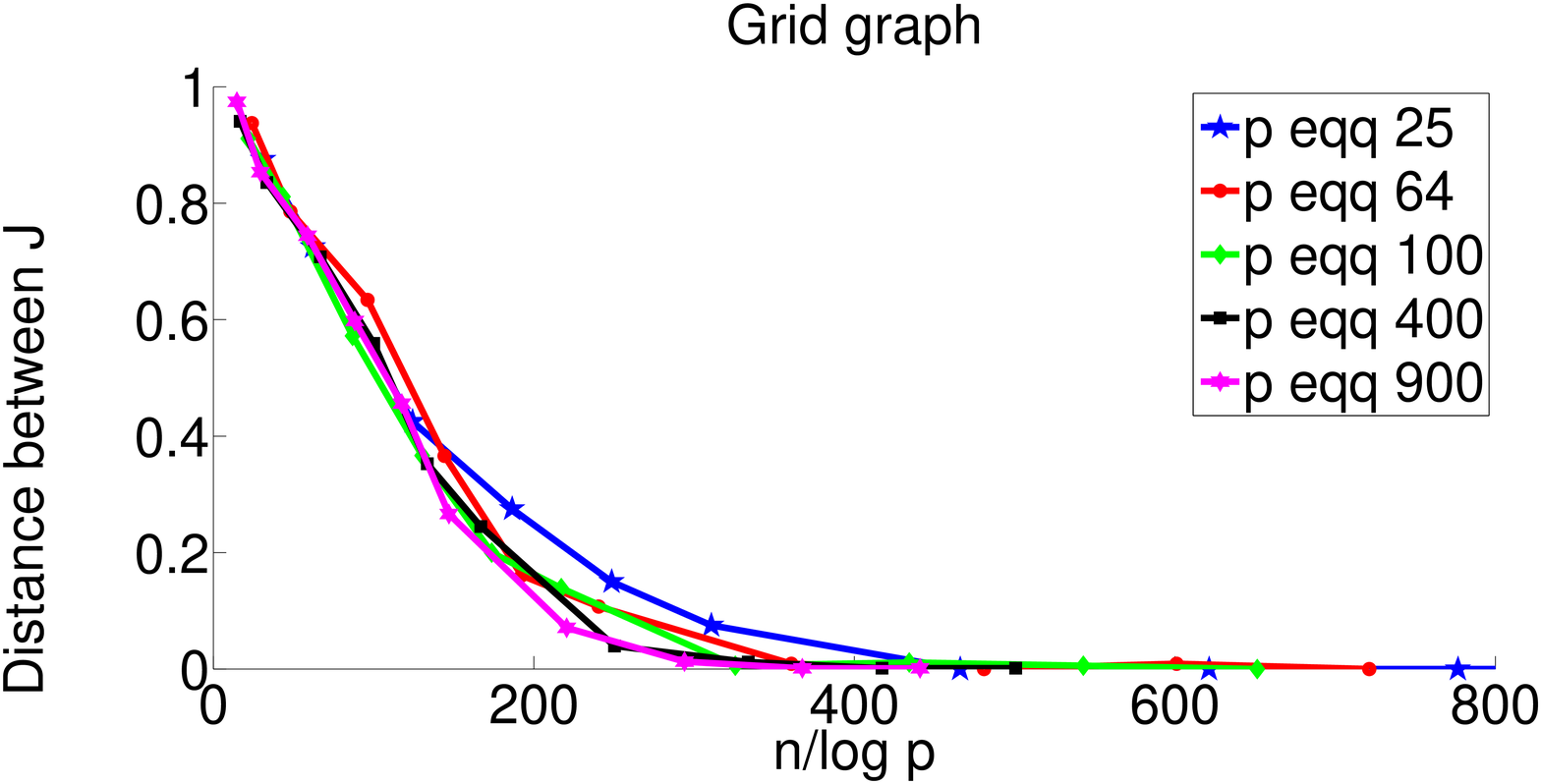}\ep\end{minipage}}
\hfil
\subfloat[b][]{\begin{minipage}{2.4in}
\bp
\psfrag{p eqq 25}[l]{\tiny $p\!=\!25$}
\psfrag{p eqq 64}[l]{\tiny $p\!=\!64$}
\psfrag{p eqq 100}[l]{\tiny $p\!=\!100$}
\psfrag{p eqq 400}[l]{\tiny $p\!=\!400$}
\psfrag{p eqq 900}[l]{\tiny $p\!=\!900$}
\psfrag{n}[l]{\scriptsize $n$}
\psfrag{Grid graph}[l]{\scriptsize Grid graph}
\psfrag{Distance between R}[l]{\scriptsize Norm. Dist\,$(\widehat{\Sigma}_R,\Sigma_R^*)$ }
\includegraphics[width=2.4in,height=1.3in]{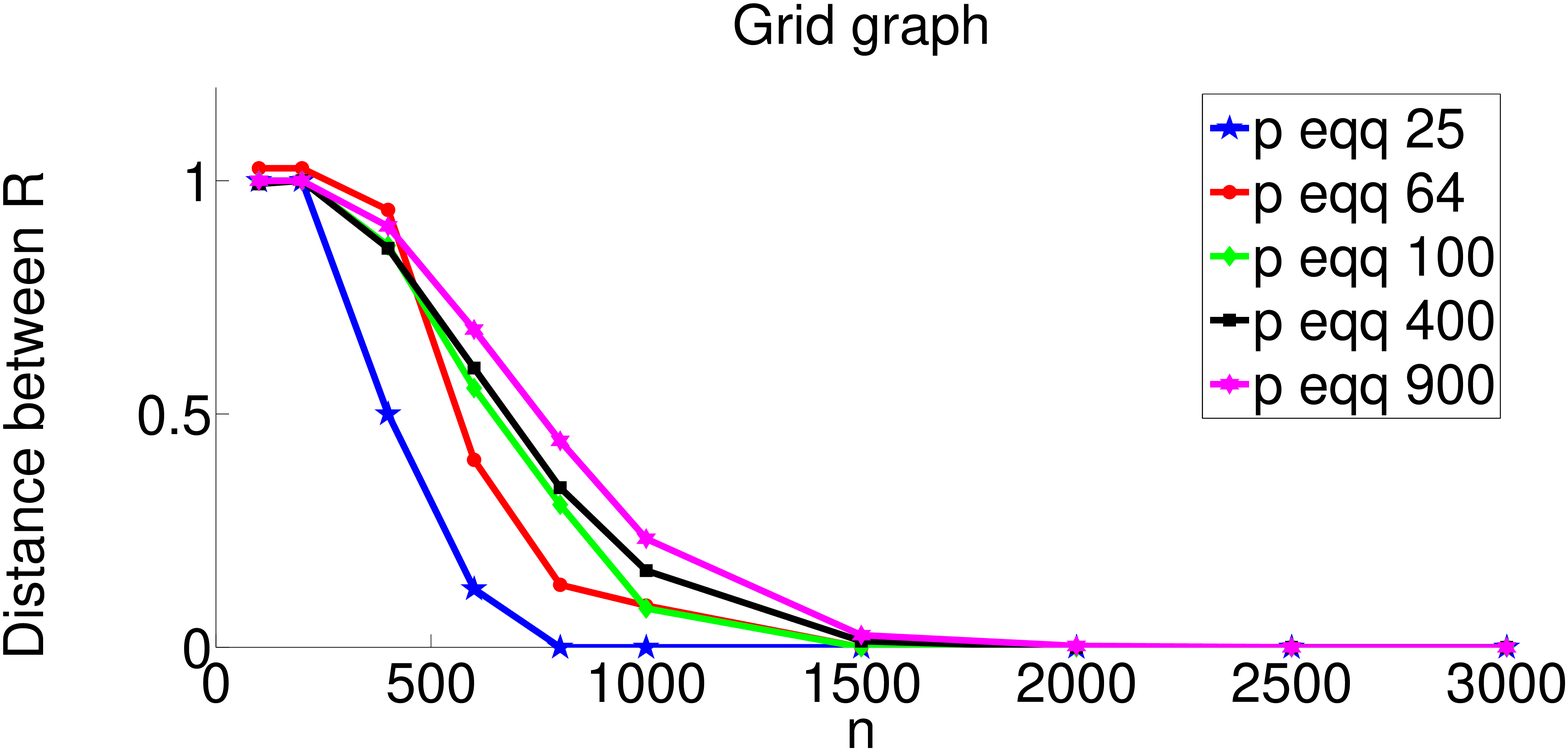}\ep\end{minipage}}
\hfil
\subfloat[b][]{\begin{minipage}{2.4in}
\bp
\psfrag{p eqq 25}[l]{\tiny $p\!=\!25$}
\psfrag{p eqq 64}[l]{\tiny $p\!=\!64$}
\psfrag{p eqq 100}[l]{\tiny $p\!=\!100$}
\psfrag{p eqq 400}[l]{\tiny $p\!=\!400$}
\psfrag{p eqq 900}[l]{\tiny $p\!=\!900$}
\psfrag{n/log p}[l]{\scriptsize $n/\log p$}
\psfrag{Grid graph}[l]{\scriptsize Grid graph}
\psfrag{Distance between R}[l]{\scriptsize Norm. Dist\,$(\widehat{\Sigma}_R,\Sigma_R^*)$ }
\includegraphics[width=2.4in,height=1.3in]{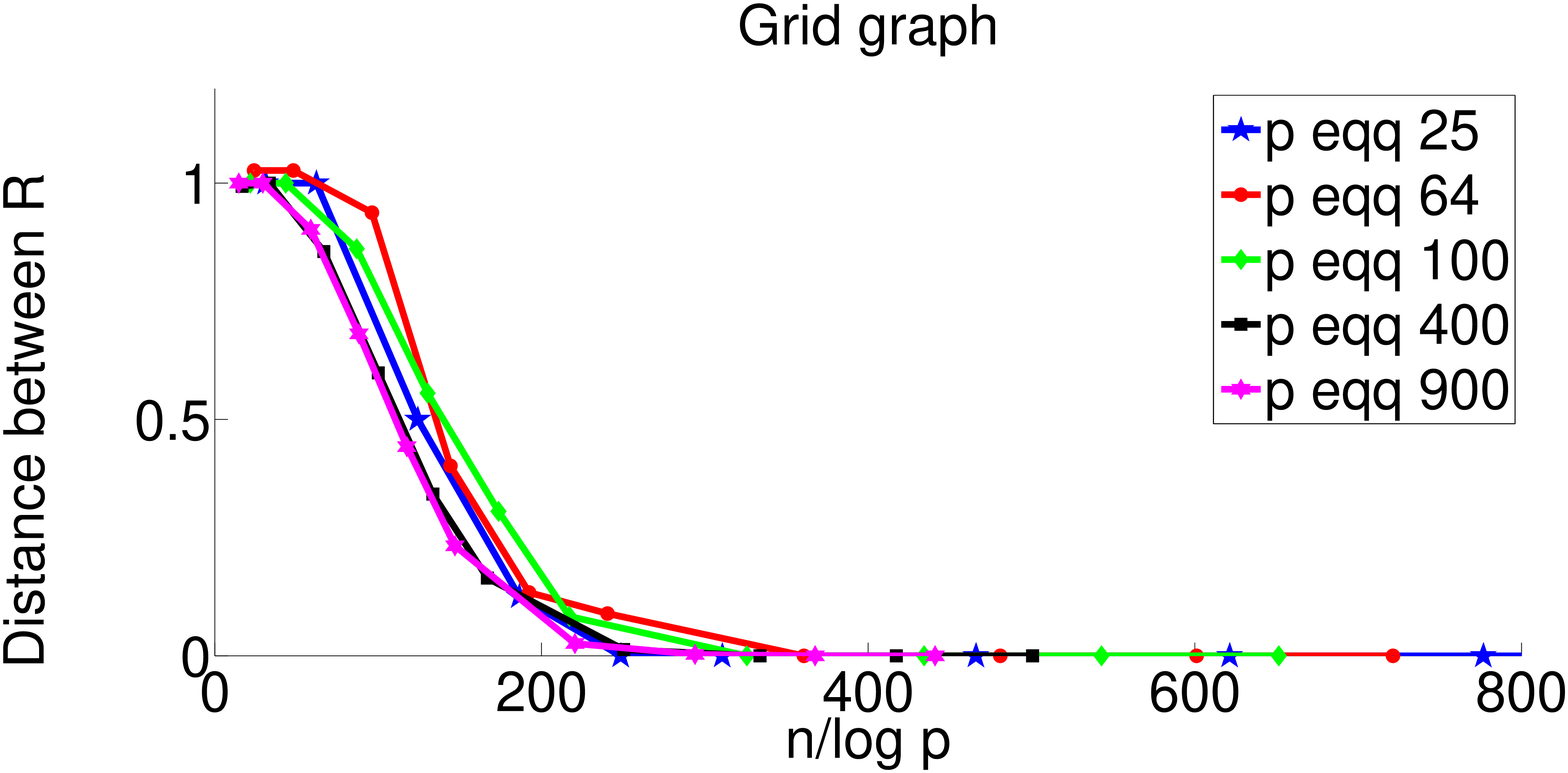}\ep\end{minipage}}
\caption{\small Simulation results for grid-structured Markov graph with different size $p$. (a-b) Normalized edit distance between the estimated Markov component $\widehat{J}_M$ and the exact Markov component $J_M^*$. In panel (b), the horizontal axis is rescaled as $n/\log p$. (c-d) Normalized edit distance between the estimated residual component $\widehat{\Sigma}_R$ and the exact residual component $\Sigma_R^*$. In panel (d), the horizontal axis is rescaled as $n/\log p$. Each point in the figures is derived  from averaging 10 trials.}\label{fig:Distance_JM_multiple_p}
\end{figure*}

We apply our method ($\ell_1 + \ell_\infty$ method) to random realizations of the above described model $\Sigma^* = {J_M^*}^{-1} - \Sigma_R^*$ with different sizes $p \in \{25,64,100,400,900\}$. Normalized $\operatorname{Dist} \left( \widehat{J}_M,J_M^* \right)$, the edit distance between the estimated and exact Markov components $\widehat{J}_M$ and $J_M^*$, and normalized $\operatorname{Dist} \left( \widehat{\Sigma}_R,\Sigma_R^* \right)$, the edit distance between the estimated and exact residual components $\widehat{\Sigma}_R$ and $\Sigma_R^*$ as a function of number of samples are plotted in Figure \ref{fig:Distance_JM_multiple_p} for different sizes $p$.\\
In Figure \ref{fig:Distance_JM_multiple_p}.a, normalized $\operatorname{Dist} \left( \widehat{J}_M,J_M^* \right)$ is plotted and in Figure \ref{fig:Distance_JM_multiple_p}.b, the same is plotted with rescaled horizontal axis $n/\log p$. We observe that by increasing the number of samples, the edit distance decreases, and by increasing the size of problem, it becomes harder to recover the components which are intuitive. More importantly, we observe in the rescaled graph that the plots for different sizes $p$ make a lineup which is consistent with the theoretical results saying that\,\footnote{Note that in the grid graph, $d=4$ is fixed for different sizes $p$.} $n=O(d^2 \log p)$ is sufficient for correct recovery. \\
Similarly, in Figure \ref{fig:Distance_JM_multiple_p}.c, normalized\,\footnote{The normalized distance for recovering residual component is greater than 1 for small $n$. Since we normalize the distance with the number of edges in the exact model, this may happen.} $\operatorname{Dist} \left( \widehat{\Sigma}_R,\Sigma_R^* \right)$ is plotted and in Figure \ref{fig:Distance_JM_multiple_p}.d, the same is plotted with rescaled horizontal axis $n/\log p$. We similarly have the initial observations that by increasing the number of samples, the edit distance decreases, and by increasing the size of problem, it becomes harder to recover the components. The theoretical sample complexity $n=O(d^2 \log p)$ is also validated in Figure \ref{fig:Distance_JM_multiple_p}.d.

The value of regularization parameters used for this simulation are provided in Table \ref{table:regularization parameters}. Since in the synthetic experiments, we know the value of $\lambda^* := \| J_M^*\|_{\infty,\operatorname{off}}$, parameter $\lambda$ is set to $\lambda^*=0.2$. It is observed that the recovery of sparsity pattern of the Markov component $J_M^*$ is fairly robust to the choice of this parameter. For choosing parameter $\gamma$, the experiment is run for several values of $\gamma$ to see which one gives the best recovery result. The effect of parameter $\gamma$ is discussed in detail in the next subsection.

\begin{table} 
\caption{Regularization parameters used for grid-structured Markov graph simulations in Figure \ref{fig:Distance_JM_multiple_p}. Note that $\gamma = c_\gamma \sqrt{\log p/n}$.} \label{table:regularization parameters}
\centering\begin{tabular}{|c|c|c|} 
\hline 
Size($p$) & $c_\gamma$ & $\lambda$ \\ 
\hline 
\hline
25 & 2.23 & 0.2 \\ 
\hline 
64 & 2.08 & 0.2 \\ 
\hline 
100 & 2.01 & 0.2 \\ 
\hline 
400 & 1.85 & 0.2 \\ 
\hline 
900 & 1.83 & 0.2 \\ 
\hline
\end{tabular} 
\end{table}

\subsubsection*{Effect of regularization parameter $\gamma$} 
\begin{figure*}[t]\centering
\subfloat[a][]{\begin{minipage}{2.4in}
\bp
\psfrag{gamma eq 1}[l]{\tiny $c_\gamma=1$}
\psfrag{gamma eq 1.3}[l]{\tiny $c_\gamma=1.3$}
\psfrag{gamma eq 2.08}[l]{\tiny $c_\gamma = 2.08$}
\psfrag{gamma eq 2.5}[l]{\tiny $c_\gamma=2.5$}
\psfrag{gamma eq 3}[l]{\tiny $c_\gamma=3$}
\psfrag{n}[l]{\scriptsize $n$}
\psfrag{Distance between J}[l]{\scriptsize Dist\,$(\widehat{J}_M,J_M^*)$ }
\includegraphics[width=2.4in,height=1.3in]{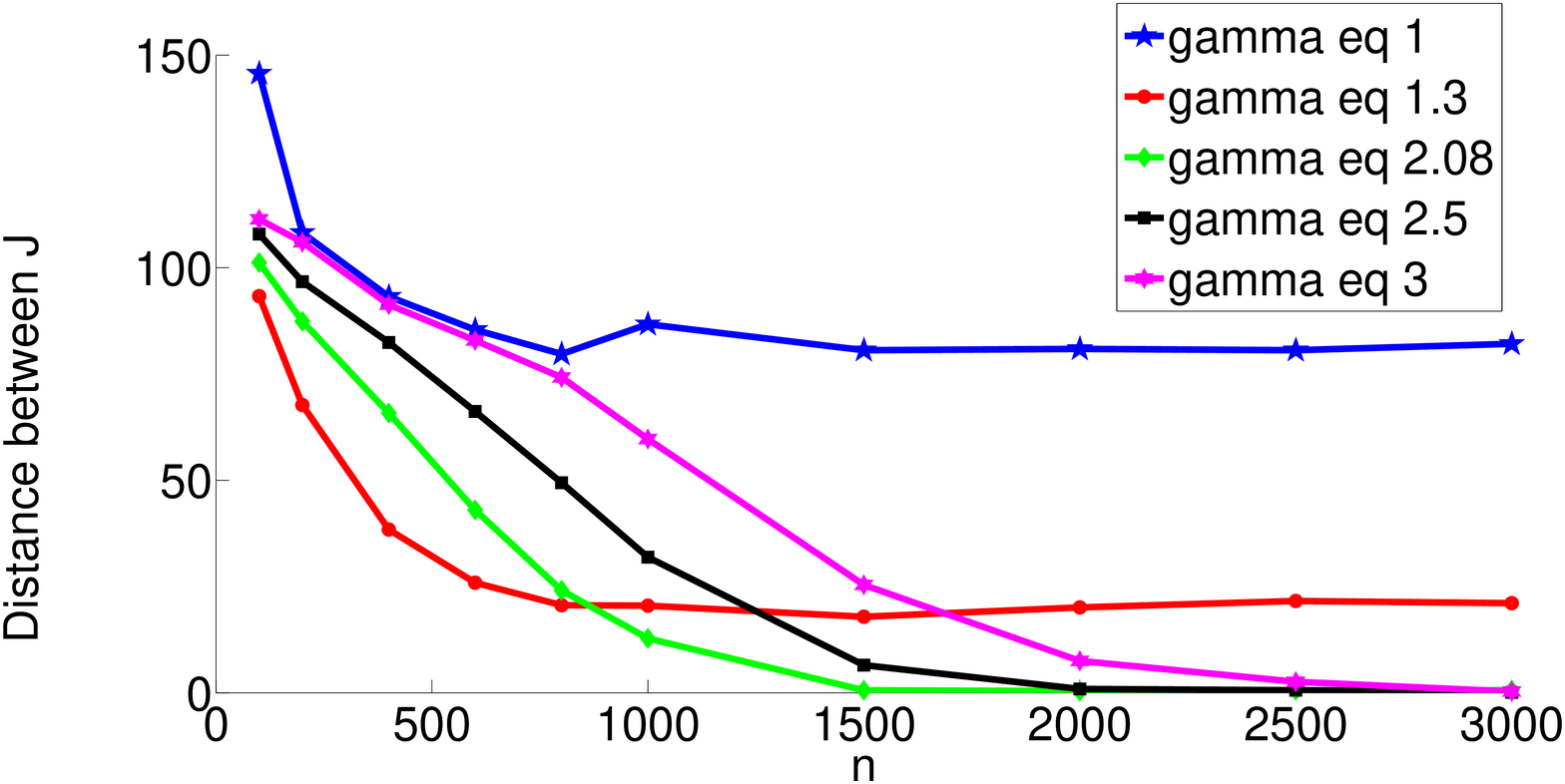}\ep
\end{minipage}}
\hfil
\subfloat[b][]{\begin{minipage}{2.4in}
\bp
\psfrag{gamma eq 1}[l]{\tiny $c_\gamma=1$}
\psfrag{gamma eq 1.3}[l]{\tiny $c_\gamma=1.3$}
\psfrag{gamma eq 2.08}[l]{\tiny $c_\gamma = 2.08$}
\psfrag{gamma eq 2.5}[l]{\tiny $c_\gamma=2.5$}
\psfrag{gamma eq 3}[l]{\tiny $c_\gamma=3$}
\psfrag{n}[l]{\scriptsize $n$}
\psfrag{Distance between R}[l]{\scriptsize Dist\,$(\widehat{\Sigma}_R,\Sigma_R^*)$ }
\includegraphics[width=2.4in,height=1.3in]{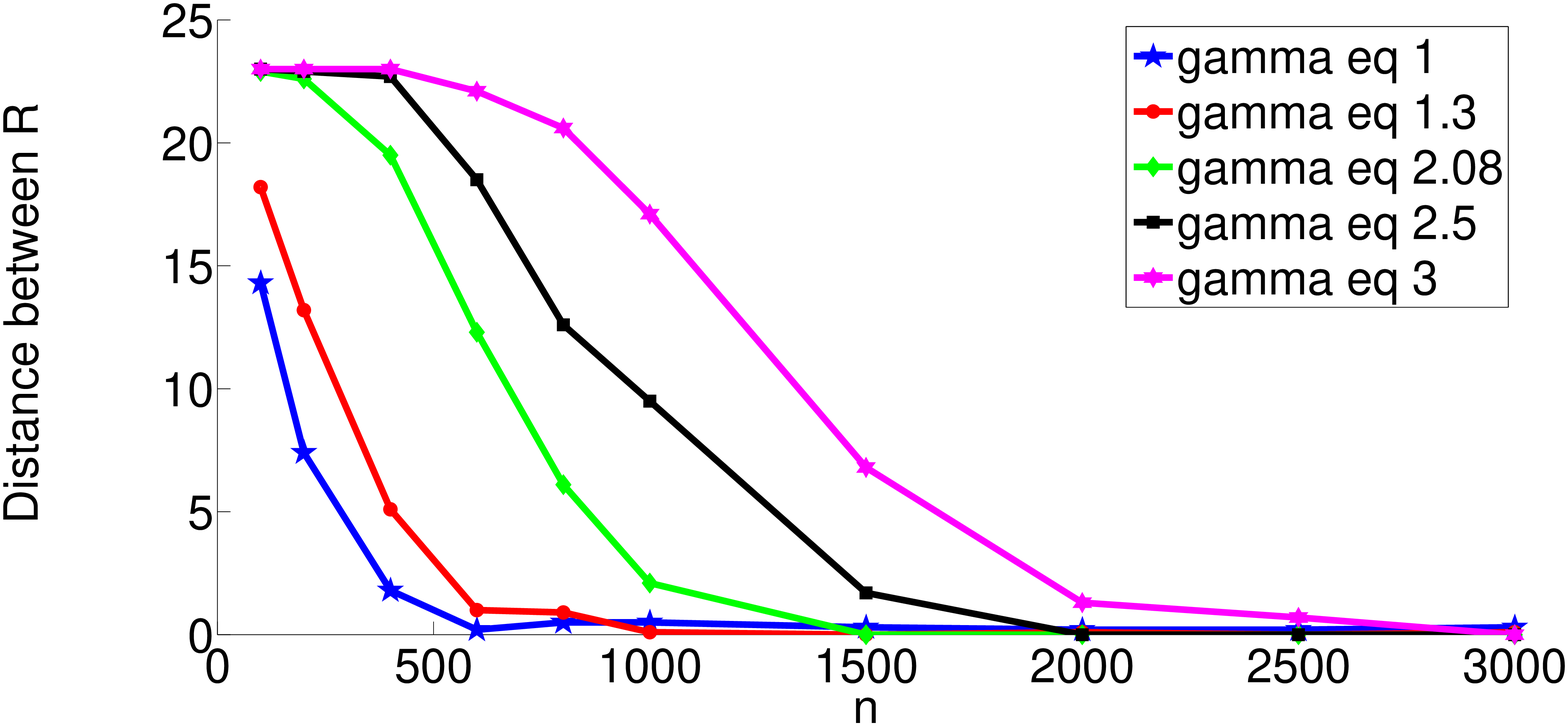}\ep\end{minipage}}
\caption{\small Simulation results for grid graph with fixed size $p=64$ and regularization parameters $\lambda=0.2$ and varying $c_\gamma \in \{1,1.3,2.08,2.5,3\}$ where $\gamma = c_\gamma \sqrt{\log p/n}$. (a) Edit distance between the estimated Markov component $\widehat{J}_M$ and the exact Markov component $J_M^*$. (b) Edit distance between the estimated residual component $\widehat{\Sigma}_R$ and the exact residual component $\Sigma_R^*$. Each point in the figures is derived  from averaging 10 trials.}\label{fig:Distance_JM_multiple_gamma}
\end{figure*}

We apply our method ($\ell_1 + \ell_\infty$ method) to random realizations of the above described grid-structured synthetic model $\Sigma^* = {J_M^*}^{-1} - \Sigma_R^*$ with fixed size $p=64$. Here, we fix the regularization parameter\,\footnote{$\lambda$ is set to the maximum absolute value of off-diagonal entries of Markov matrix $J_M^*$.} $\lambda=0.2$ and change the regularization parameter $\gamma = c_\gamma \sqrt{\log p/n}$ where $c_\gamma \in \{1,1.3,2.08,2.5,3\}$. The edit distance between the estimated and exact Markov components $\widehat{J}_M$ and $J_M^*$, and the edit distance between the estimated and exact residual components $\widehat{\Sigma}_R$ and $\Sigma_R^*$ are plotted in Figure \ref{fig:Distance_JM_multiple_gamma}. We observe the pattern that for $c_\gamma$ less than some optimal value $c_\gamma^*$, the Markov component is not recovered, and for values greater than the optimal value, the components are recovered with different statistical efficiency, where by increasing $c_\gamma$, the statistical rate of Markov component recovery becomes worse. For the simulations of previous subsection provided in Figure \ref{fig:Distance_JM_multiple_p}, we choose some regularization parameter close to $c_\gamma^*$. For example, we choose $c_\gamma=2.08$ for $p=64$ as suggested by Figure \ref{fig:Distance_JM_multiple_gamma}.

\subsubsection*{Comparing $\ell_1 + \ell_\infty$ and $\ell_1$ methods}
\begin{figure*}[t]\centering
\subfloat[a][]{\begin{minipage}{2.4in} \label{{fig:Distance_JM}}
\bp\psfrag{a1111111111111111111111}[l]{\scriptsize $\ell_1+\ell_\infty$ method}
\psfrag{a2}[l]{\scriptsize $\ell_1$ method}
\psfrag{n}[l]{\scriptsize $n$}
\psfrag{Distance between J}[l]{\scriptsize Dist\,$(\widehat{J}_M,J_M^*)$ }
\includegraphics[width=2.4in,height=1.3in]{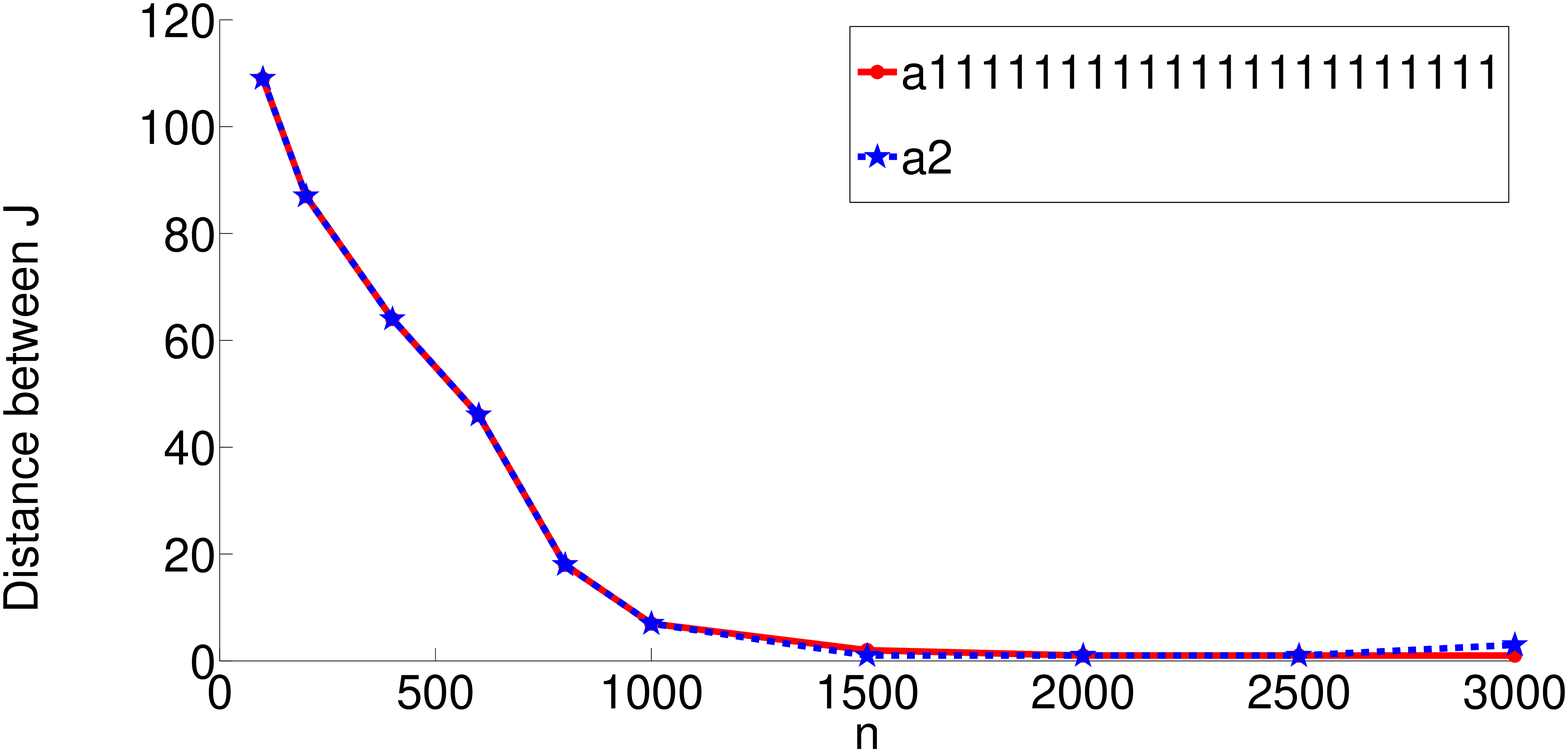}\ep\end{minipage}}
\hfil
\subfloat[b][]{\begin{minipage}{2.4in}
\bp\psfrag{a1111111111111111111111}[l]{\scriptsize $\ell_1+\ell_\infty$ method}
\psfrag{a2}[l]{\scriptsize $\ell_1$ method}
\psfrag{n}[l]{\scriptsize $n$}
\psfrag{Distance between R}[l]{\scriptsize Dist\,$(\widehat{\Sigma}_R,\Sigma_R^*)$ }
\includegraphics[width=2.4in,height=1.3in]{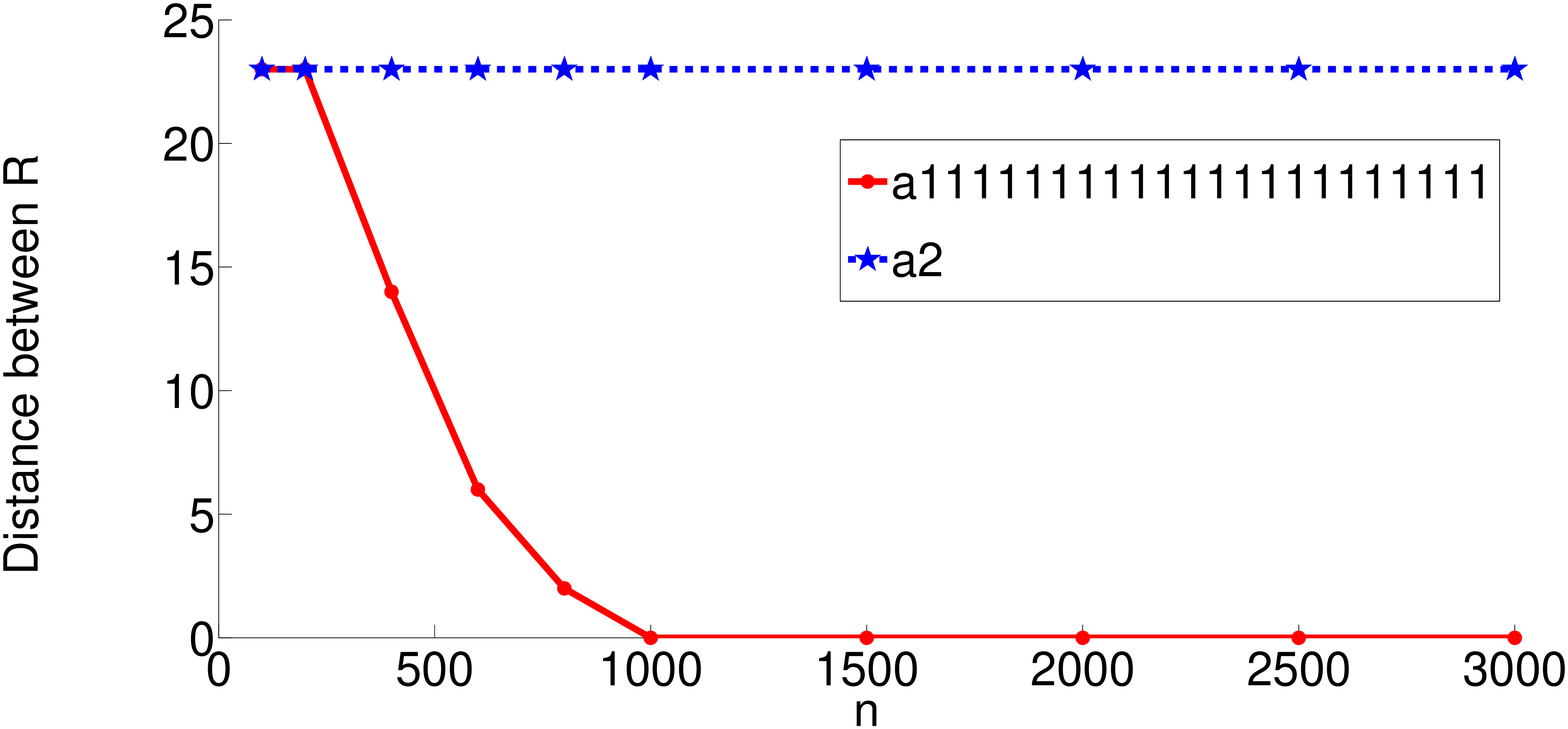}\ep\end{minipage}
}
\hfil
\subfloat[c][]{\begin{minipage}{2.4in}
\bp\psfrag{a1111111111111111111111}[l]{\scriptsize $\ell_1+\ell_\infty$ method}
\psfrag{a2}[l]{\scriptsize $\ell_1$ method}
\psfrag{n}[l]{\scriptsize $n$}
\psfrag{J error}[l]{\scriptsize $\| J^* - \widehat{J} \|_\infty$ }
\includegraphics[width=2.4in,height=1.3in]{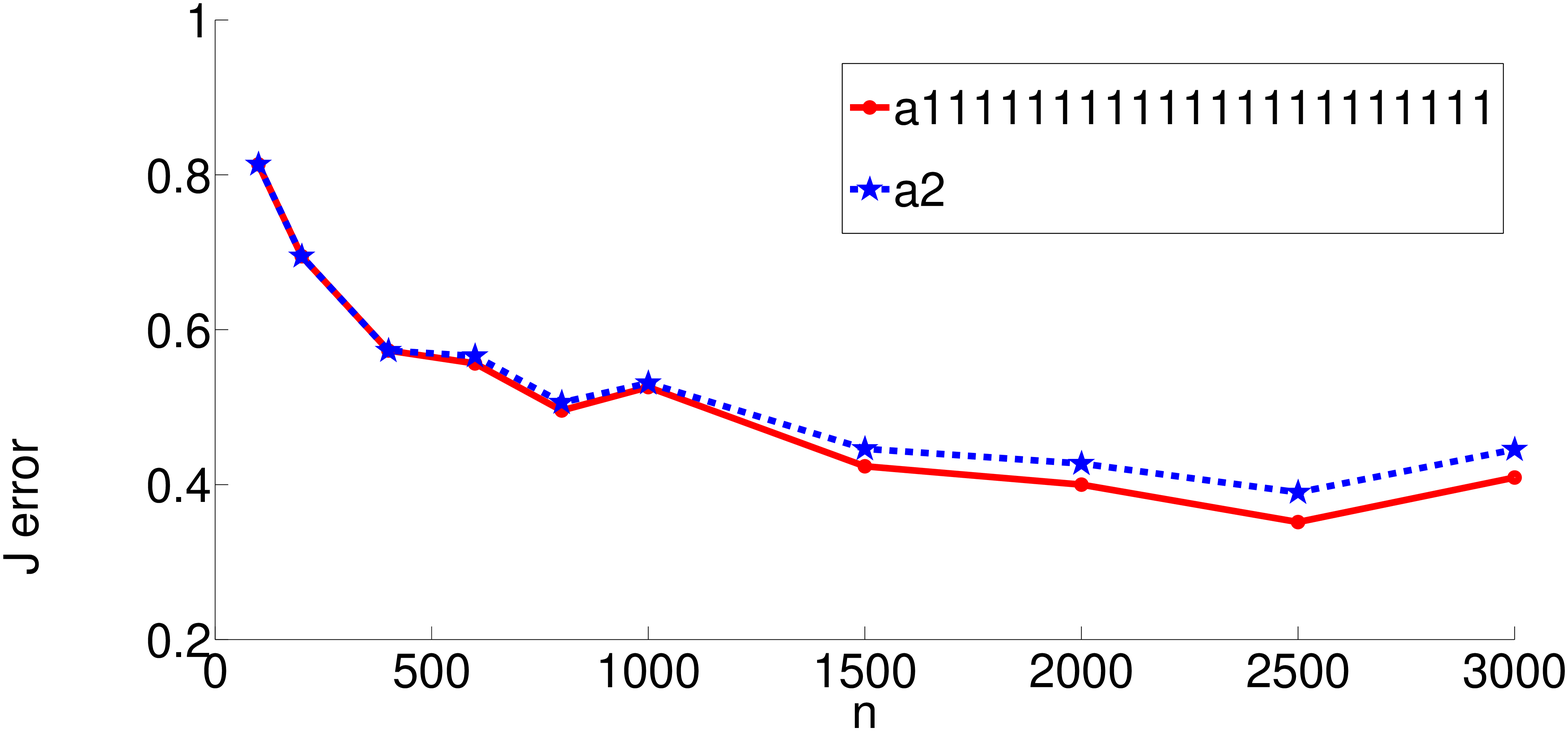}\ep\end{minipage}}
\caption{\small Simulation results for grid graph with size $p=64$. (a) Edit distance between the estimated Markov component $\widehat{J}_M$ and the exact Markov component $J_M^*$. (b) Edit distance between the estimated residual component $\widehat{\Sigma}_R$ and the exact residual component $\Sigma_R^*$. (c) Precision matrix estimation error $\| J^* - \widehat{J} \|_\infty$ , where $\hJ= \hJ_M$ for $\ell_1$ method and $\hJ= \bigl( \hJ_M^{-1} - \hSigma_R \bigr) ^{-1}$ for $\ell_1+\ell_\infty$ method.
}\label{fig:Distance_JM}
\end{figure*}

We apply $\ell_1 + \ell_\infty$ and $\ell_1$ methods to a random realization of the above described grid-structured synthetic model\,\footnote{Here, we choose the nonzero off-diagonal entries of $J_M^*$ randomly from $\{-0.2,0.2\}$.} $\Sigma^* = {J_M^*}^{-1} - \Sigma_R^*$ with size $p=64$. The edit distance between the estimated and exact Markov components $\widehat{J}_M$ and $J_M^*$ is plotted in Figure \ref{fig:Distance_JM}.a. We observe that the behaviour of $\ell_1 + \ell_\infty$ method is very close to $\ell_1$ method which suggests that sparsity pattern of $J_M^*$ can be estimated efficiently under either methods. 
The edit distance between the estimated and exact residual components $\widehat{\Sigma}_R$ and $\Sigma_R^*$ is plotted in Figure \ref{fig:Distance_JM}.b. Since there is not any off-diagonal $\ell_\infty$ constraints in $\ell_1$ method, it can not recover the residual matrix $\Sigma_R^*$. Finally the $\ell_\infty$-elementwise norm of error between the estimated precision matrix $\hJ$ and the exact precision matrix $J^*$ is sketched for both methods in Figure \ref{fig:Distance_JM}.c. We observe the advantage of proposed $\ell_1 + \ell_\infty$ method in estimating the overall model precision matrix $J^* = {\Sigma^*}^{-1}$. Note that the same regularization parameters provided in Table \ref{table:regularization parameters} are used for the simulations of this subsection, except for $\ell_1$ method that we have $\lambda=\infty$.

\subsubsection*{Benefit of applying LBP\,\footnote{Loopy Belief Propagation} to the proposed model}
\begin{figure*}[t]\centering
\subfloat[(a)][]{\begin{minipage}{2.4in}
\bp\psfrag{J11111111111111111111111111111}[l]{\scriptsize LBP applied to $J^*$ model}
\psfrag{JM}[l]{\scriptsize LBP applied to $J_M^*$ model}
\psfrag{iteration}[l]{\small iteration}
\psfrag{average mean error}[l]{\small average mean error}
\includegraphics[width=2.4in,height=1.3in]{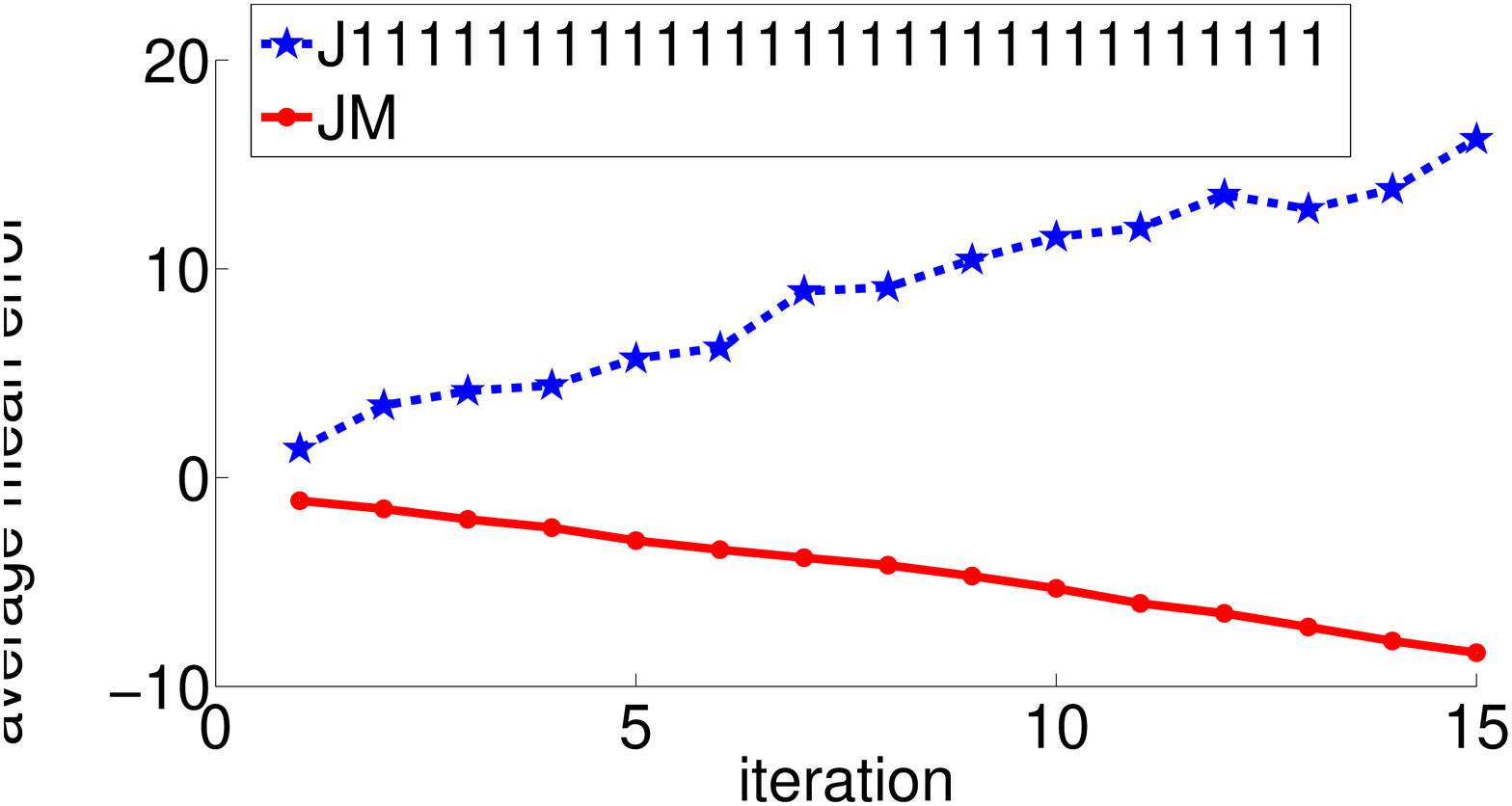}\ep\end{minipage}}
\hfil
\subfloat[(b)][]{\begin{minipage}{2.4in}
\bp\psfrag{J11111111111111111111111111111}[l]{\scriptsize LBP applied to $J^*$ model}
\psfrag{JM}[l]{\scriptsize LBP applied to $J_M^*$ model}
\psfrag{iteration}[l]{\small iteration}
\psfrag{average variance error}[l]{\small average variance error}
\includegraphics[width=2.4in,height=1.3in]{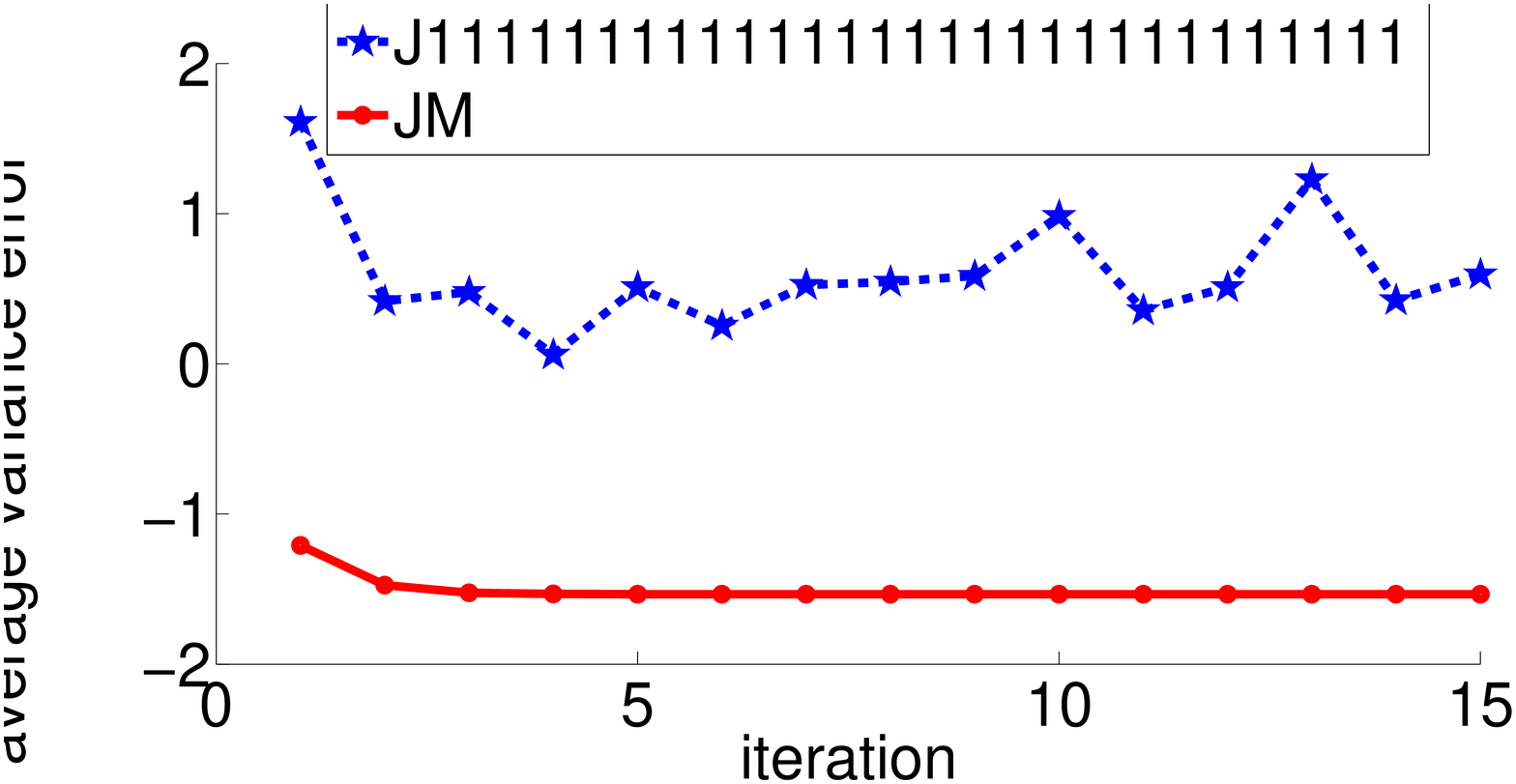}\ep\end{minipage}}
\caption{\small Performance under loopy belief propagation for the overall model ($J^*$) and the Markov component ($J_M^*$).}\label{fig:inference_NOTconverged2}
\end{figure*}

We compare the result of applying LBP to $J^*$ and $J_M^*$ components of a random realization of the above described grid-structured synthetic model\,\footnote{Here, we choose 0.5 fraction of Markov edges randomly to introduce residual edges.}. The log of average mean and variance errors over all nodes are sketched in Figure \ref{fig:inference_NOTconverged2} throughout the iterations.
We observe that LBP does not converge for $J^*$ model. It is shown in \cite{Malioutov&etal:06JMRL} that if a model is walk-summable, then the mean estimates under LBP converge and are correct. The spectral norms of the partial correlation matrices are $\gennorm{ \overline{R}_M } = 0.8613$ and $\gennorm{ \overline{R} } = 3.2446$ for $J_M^*$ and $J^*$ models respectively. Thus, the matrix $J^*$ is not walk-summable and therefore its convergence under LBP is not guaranteed and this is seen in Figure \ref{fig:inference_NOTconverged2}. On the other hand, LBP is accurate for $J^*_M$ matrix. Thus, our method learns models which are better suited for inference under loopy belief propagation.

\subsection{Real Data}

The proposed algorithm is also applied to foreign exchange rate and monthly stock returns data sets to learn a Markov plus residual model introduced in the paper. It is important to note that the real data sets can be modeled by different models not necessarily satisfying the conditions proposed in this paper. But, here we observe that the resulting Markov plus residual models are fairly interpretable for the corresponding real data sets. The interpretations are discussed in detail in the following sections.

\subsubsection*{Foreign Exchange Rate Data} 
\begin{figure}\centering{
\bp
\includegraphics[width=4in]{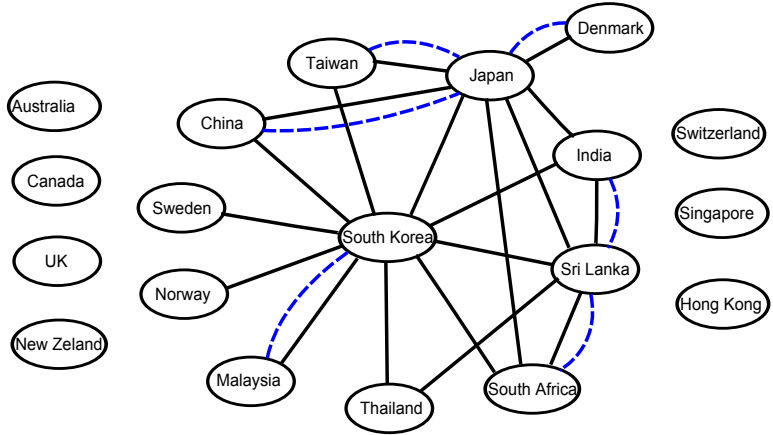}\ep}
\caption{\small Markov and independence graph structures for the foreign exchange rate data set with regularization parameters $\gamma = 20$ and $\lambda = 0.004$. Solid edges indicate Markov model and dotted edges indicate independence model.}\label{fig:exchange rate}
\end{figure}

In this section, we apply the proposed algorithm to the foreign exchange rate data set\footnote{Dataset available at http://research.stlouisfed.org/fred2/categories/15/downloaddata}. The dataset includes monthly exchange rates of 19 countries currency with respect to US dollars from October 1983 to January 2012. Thus, the dataset has 340 samples of 19 variables. We apply the optimization program \eqref{convex_prog_dual_sample_case} with a slight modification. Since the underlying model for this data set does not necessarily satisfy the proposed eigenvalue condition (A.6), we need to make sure that the overall covariance matrix estimation $\widehat{\Sigma}$ is positive definite and thus a valid covariance matrix. We add an additional constraint to the optimization program \eqref{convex_prog_dual_sample_case}, imposing a lower bound on the minimum eigenvalue of overall covariance matrix $\lambda_{min} (\Sigma)$, i.e., $\lambda_{min} (\Sigma) \geq \sigma_{min}$. The parameter $\sigma_{min}$ is set to 0.001 in this experiment. \\
The resulting edges of Markov and residual matrices for some moderate choice of regularization parameters $\gamma = 20$ and $\lambda = 0.004$ are plotted in Figure \ref{fig:exchange rate}. The choice of regularization parameters are further discussed at the end of this subsection. We observe sparsity on both Markov and residual structures.
There are two main observations in the learned model in Figure \ref{fig:exchange rate}. First, it is seen that the statistical dependencies of foreign exchange rates are correlated with the geographical locations of countries, e.g., it is observed in the learned model that the exchange rates of Asian countries are more correlated. We can refer to Asian countries ``South Korea", ``Japan",``China",``Sri Lanka", ``Taiwan", ``Thailand" and ``India" in the Markov model where several edges exist between them while other nodes in the graph have much lower degrees. We observe similar patterns in the residual matrix, e.g., there is an edge between ``India" and ``Sri Lanka" in the residual model.
We also see the interesting phenomena in the Markov graph that there exist some high degree nodes such as ``South Korea" and ``Japan". The presence of high degree nodes suggests that incorporating hidden variables can further lead to sparser representations, and this has been observed before in other works, e.g., \cite{Choi&etal:10SP}, \cite{Chandrasekaran:10latent} and \cite{Choi&etal:10JMLR}.

The regularization parameters are chosen such that the resulting Markov and residual graphs are reasonably sparse, while still being informative. Increasing the parameter $\gamma$ makes both Markov and residual components sparser, and increasing parameter $\lambda$ makes the residual component sparser. In addition, it is worth discussing the fact that we chose parameter $\gamma$ relatively large compared to parameter $\lambda$ in this simulation. In Theorem 4, we have $\gamma = C_1 \sqrt{\log p / n}$ and $\lambda = \lambda^* + C_2 \sqrt{\log p / n}$. Now, if $C_1$ is large compared to $C_2$ and furthermore $\lambda^*$ is small, $\gamma$ can be larger than $\lambda$. Hence, we have an agreement between theory and practice.

\subsubsection*{Monthly Stock Returns Data} 
\begin{figure}\centering{
\bp
\includegraphics[width=4in]{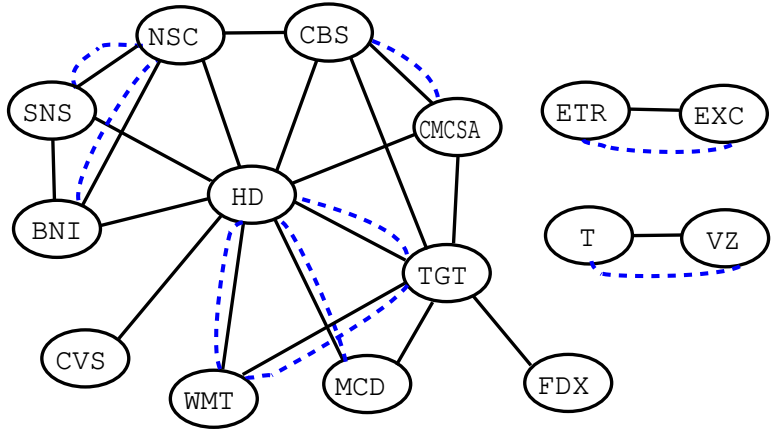}\ep}
\caption{\small Markov and independence graph structures for the monthly stock returns data set with regularization parameters $\gamma = 2.2e-03$ and $\lambda = 1e-04$. Solid edges indicate Markov model and dotted edges indicate independence model.}\label{fig:stock_Market_Markov}
\end{figure}

In this section we apply the algorithm to monthly stock returns of a number of companies in the S\&P 100 stock index. We pick 17 companies in divisions ``E.Trans, Comm, Elec\&Gas" and ``G.Retail Trade" and apply the optimization program \eqref{convex_prog_primal_sample_case} to their stock returns data to learn the model. The resulting edges for Markov and residual matrices are plotted in Figure \ref{fig:stock_Market_Markov} for regularization parameters $\gamma = 2.2e-03$ and $\lambda = 1e-04$. There is sparsity on both Markov and residual structure. The isolated nodes  in the Markov graph are not presented in the figure. We see in both Markov and residual graphs  that there exist higher correlations among stock returns of companies in the same division or industry. There are 5 connected partitions in the residual graph. e.g. nodes ``HD", ``WMT", ``TGT" and ``MCD", all belonging to division Retail Trade form  a partition. This is also observed for the  telecommunication industries (companies ``T" and ``VZ") and energy industries (companies ``ETR" and ``EXC"). We see a similar pattern in the Markov graph but with more edges. Similar to exchange rate data set results, we also observe high degree nodes in the Markov graph such as ``HD" and ``TGT" which suggest incorporating hidden nodes.

\section{Conclusion}

In this paper, we provided an in-depth study of convex optimization methods and guarantees for high-dimensional covariance matrix decomposition.
Our methods unify the existing results for sparse covariance/precision estimation and introduce a richer class of models with sparsity in multiple domains.
We provide consistency guarantees for estimation in both the Markov and the  residual domains, and establish efficient sample complexity results for our method. These findings open up many future directions to explore. One important aspect is to relax the sparsity constraints imposed in the two domains, and to develop new methods to enable decomposition of such models. Other considerations include extension to   discrete models and other models for the residual covariance matrix (e.g. low rank matrices). Such findings will push the envelope of efficient models for high-dimensional estimation.
It is worth mentioning while in many scenarios it is important to incorporate latent variables, in our framework it is challenging to incorporate both latent variables as well as marginal independencies, and provide learning guarantees, and we defer it to future work.

\subsection*{Acknowledgements}
We thank Karthik Mohan for helpful discussions on running experiments.
We also acknowledge useful discussions with Max Welling, Babak Hassibi and Martin Wainwright. We also thank Bin Yu and the JMLR reviewers for valuable comments that have significantly improved the manuscript. 
M. Janzamin is supported by  NSF Award CCF-1219234 and ARO Award W911NF-12-1-0404. A. Anandkumar is supported in part by  Microsoft Faculty Fellowship, NSF Career award CCF-1254106, NSF Award CCF-1219234, AFOSR Award FA9550-10-1-0310, and ARO Award W911NF-12-1-0404.

\renewcommand{\appendixpagename}{Appendix}

\appendixpage

\appendix

\section{Duality Between Programs} \label{appendix:duality}
In this section we prove duality between programs \eqref{convex_prog_primal_sample_case} and \eqref{convex_prog_dual_sample_case} (when the positive-definiteness constraint $\Sigma_M - \Sigma_R \succ 0$ is dropped). By doing this, the duality between programs \eqref{convex_prog_primal} and \eqref{convex_prog_dual} is also proved since they are special cases of \eqref{convex_prog_primal_sample_case} and \eqref{convex_prog_dual_sample_case} when $\gamma$ is set to zero and $\widehat{\Sigma}^n$ is substituted with $\Sigma^*$.

Before we prove duality, we introduce the concept of \textit{subdifferential} or \textit{subgradient} for a convex function not necessarily differentiable. Subgradient (subdifferential) generalizes the gradient (derivative) concept to nondifferentiable functions. Supposing convex function $f: \mathbb{R}^n \rightarrow \mathbb{R}$, the subgradient at a point $x_0$ which is usually denoted by $\partial f(x_0)$ consists of all vectors $c$ such that
\begin{equation}
f(x) \geq f(x_0) + \langle c,x-x_0 \rangle, \quad \forall x \in \operatorname{Dom} f. 
\end{equation}

In order to prove duality, we start from program \eqref{convex_prog_dual_sample_case} (when the positive-definiteness constraint $\Sigma_M - \Sigma_R \succ 0$ is dropped) and derive the primal form \eqref{convex_prog_primal_sample_case}. Program \eqref{convex_prog_dual_sample_case} can be written in the following equivalent form where $\lambda_1$ goes to infinity and $\lambda_2$ is used instead of $\lambda$.
\begin{align}
\label{convex_prog_dual_generalized}
\bigl( \widehat{\Sigma}_M, \widehat{\Sigma}_R \bigr) := \argmax_{\Sigma_M \succ 0, \Sigma_R} & \log \det \Sigma_M - \lambda_1 \| \Sigma_R \| _{1, \operatorname{on}} - \lambda_2 \| \Sigma_R \| _{1, \operatorname{off}} \nonumber \\
\operatorname{s.t.} \ \ & \ \|\widehat{\Sigma}^n - \Sigma_M + \Sigma_R\|_{\infty, \operatorname{off}} \leq \gamma   , \\
& \ \bigl( \Sigma_M \bigr)_d - \bigl( \Sigma_R \bigr)_d = \bigl( \widehat{\Sigma}^n \bigr)_d . \nonumber
\end{align}
By introducing the dual variable $J_M$ for above program, we have:
\begin{equation}
\min_{\substack{\|J_M\|_{\infty, \operatorname{on}} \leq \lambda_1 \\ \|J_M\|_{\infty, \operatorname{off}} \leq \lambda_2}} - \langle J_M,\Sigma_R \rangle = - \lambda_1 \|\Sigma_R\|_{1, \operatorname{on}} - \lambda_2 \|\Sigma_R\|_{1, \operatorname{off}} ,
\end{equation}
where $(\widehat{J}_M)_{\operatorname{on}} \in \lambda_1 \partial \|\widehat{\Sigma}_R\|_{1, \operatorname{on}}$, $(\widehat{J}_M)_{\operatorname{off}} \in \lambda_2 \partial \|\widehat{\Sigma}_R\|_{1, \operatorname{off}}$ minimizes the above program. Thus, we have the following equivalent form for program \eqref{convex_prog_dual_generalized}:
\begin{equation}
\min_{\substack{\|J_M\|_{\infty, \operatorname{on}} \leq \lambda_1 \\ \|J_M\|_{\infty, \operatorname{off}} \leq \lambda_2}} \ \ \max_{\substack{\Sigma_M \succ 0, \Sigma_R \\ \|\widehat{\Sigma}^n - \Sigma_M + \Sigma_R\|_{\infty , \operatorname{off}} \leq \gamma \\ \left( \Sigma_M \right)_d - \left (\Sigma_R \right)_d = \left( \widehat{\Sigma}^n \right)_d}} \log \det \Sigma_M - \langle J_M,\Sigma_R \rangle ,
\end{equation}
where the order of programs is exchanged. If we define the new variable $\Sigma = \Sigma_M - \Sigma_R$, and use $\Sigma$ as the new variable in the program instead of $\Sigma_R$, the inner max program becomes
\begin{equation}
\max_{\substack{\Sigma_M \succ 0 , \Sigma \\ \|\widehat{\Sigma}^n - \Sigma\|_{\infty , \operatorname{off}} \leq \gamma, \Sigma_d = \left( \widehat{\Sigma}^n \right)_d}} \log \det \Sigma_M - \langle  J_M,\Sigma_M \rangle + \langle J_M,\Sigma \rangle .
\end{equation}
Since the objective function and constraints are disjoint functions of variables $\Sigma$ and $\Sigma_M$, we can do optimization individually for two variables. The optimizers are $\widehat{\Sigma}_M = J_M^{-1}$ and $\widehat{\Sigma} = \widehat{\Sigma}^n + \gamma Z_{\gamma}$, where $Z_{\gamma}$ is a member of the subgradient of $\|\cdot\|_{1,\operatorname{off}}$ evaluated at point $J_M$, i.e.,
\begin{equation}
(Z_{\gamma})_{ij} = \left\{ \begin{array}{lcl}
		0 & \operatorname{for} & i=j \\
		\in [-1,1] & \operatorname{for} & i\neq j, \bigl( J_M \bigr)_{ij}=0 \\
		\sign \bigl( \bigl( J_M \bigr)_{ij} \bigr) & \operatorname{for} & i\neq j, \bigl( J_M \bigr)_{ij} \neq 0.
		\end{array}\right.
\end{equation}
Also note that since $\Sigma_M$ should be positive definite, the variable $J_M$ should be also positive definite. Therefore, it adds another constraint $J_M \succ 0$. If we substitute these optimizers, we get the dual program
\begin{equation}
\min_{\substack{J_M \succ 0 \\ \|J_M\|_{\infty, \operatorname{on}} \leq \lambda_1 \\ \|J_M\|_{\infty, \operatorname{off}} \leq \lambda_2}} \langle \widehat{\Sigma}^n,J_M \rangle - \log \det J_M + \gamma \|J_M\|_{1, \operatorname{off}},
\end{equation}
which is equivalent to \eqref{convex_prog_primal_sample_case} when $\lambda_1$ goes to infinity and therefore the result is proved.

\section{Characterization of the Proposed Optimization Programs} \label{appendix: optimal solution characterization}

We proposed programs \eqref{convex_prog_primal} and \eqref{convex_prog_primal_sample_case} to do decomposition and estimation respectively. Former is used to decompose exact statistics to its Markov and residual covariance components and the latter is used to estimate decomposition components given sample covariance matrix. In this appendix we characterize optimal solutions of these optimization programs. Both programs are convex and therefore the optimal solutions can be characterized using standard convex optimization theory. Note that the proof of following lemmas is mentioned after the remarks.

\begin{lem}
\label{optimal_solution_characterization}

For any $\lambda > 0$, primal problem \eqref{convex_prog_primal} has a \textit{unique} solution $\widehat{J}_M \succ 0$ which is characterized by the following equation:
\begin{equation}
\Sigma^* - \widehat{J}_M^{-1} + \widehat{Z} = 0,
\label{Lagrangian optimality condition}
\end{equation}
where $\widehat{Z}$ has the following form
\begin{equation}
\widehat{Z}_{ij}= \left\{ \begin{array}{lcl}
		0 & \operatorname{for} & i = j \\
		0 & \operatorname{for} & i \neq j, |\bigl( \widehat{J}_M \bigr)_{ij}| < \lambda \\
		\widehat{\alpha}_{ij} \sign \bigl( \bigl( \widehat{J}_M \bigr)_{ij} \bigr) & \operatorname{for} & i \neq j, |\bigl( \widehat{J}_M \bigr)_{ij}| = \lambda,
		\end{array}\right.
\label{Z_matrix_original}
\end{equation}
in which $\widehat{\alpha}_{ij}$ can only take nonnegative values, i.e., we have $\widehat{\alpha}_{ij} \geq 0$.
\end{lem}


\textbf{Remark}: Comparing Lagrangian optimality condition in \eqref{Lagrangian optimality condition} with relation $\Sigma^* = \widehat{J}_M^{-1} - \widehat{\Sigma}_R$ between solutions of primal-dual optimization programs (derived in Appendix \ref{appendix:duality}) implies the equality $\widehat{\Sigma}_R = \widehat{Z}$. Thus, $\widehat{\Sigma}_R$ entries are determined by Lagrangian multipliers of primal program. More specifically, we have
\begin{equation}
(\widehat{\Sigma}_R)_{ij}= \left\{ \begin{array}{lcl}
		0 & \operatorname{for} & i = j \\
		0 & \operatorname{for} & i \neq j, |\bigl( \widehat{J}_M \bigr)_{ij}| < \lambda \\
		\widehat{\alpha}_{ij} \sign \bigl( \bigl( \widehat{J}_M \bigr)_{ij} \bigr) & \operatorname{for} & i \neq j, |\bigl( \widehat{J}_M \bigr)_{ij}| = \lambda,
		\end{array}\right.
\label{SigmaR_JM_relation}
\end{equation}
where $\widehat{\alpha}_{ij} \geq 0$ are the Lagrangian multipliers of primal program \eqref{convex_prog_primal}. \\

\begin{lem}
\label{optimal_solution_characterization_sample_version}

For any $\lambda > 0$, $\gamma \geq 0$ and sample covariance matrix $\widehat{\Sigma}^n$ with strictly positive diagonal entries, primal problem \eqref{convex_prog_primal_sample_case} has a \textit{unique} solution $\widehat{J}_M \succ 0$ which is characterized by the equation
\begin{equation}
\widehat{\Sigma}^n - \widehat{J}_M^{-1} + \widehat{Z} = 0,
\label{Lagrangian optimality condition sample version}
\end{equation}
where $\widehat{Z} = \widehat{Z}_{\alpha} + \gamma \widehat{Z}_{\gamma}$. Matrix $\widehat{Z}_{\gamma} \in \partial \|\widehat{J}_M\|_{1, \operatorname{off}}$ and $\widehat{Z}_{\alpha}$ is represented as in \eqref{Z_matrix_original} for some Lagrangian multipliers $\widehat{\alpha}_{ij} \geq 0$.
%
%
\end{lem}


\textbf{Remark}: Comparing Lagrangian optimality condition in \eqref{Lagrangian optimality condition sample version} with relation $\widehat{\Sigma}^n = \widehat{J}_M^{-1} - \widehat{\Sigma}_R - \gamma \widehat{Z}_{\gamma}$ between solutions of primal-dual optimization programs (derived in Appendix \ref{appendix:duality}) implies the equality $\widehat{\Sigma}_R = \widehat{Z}_{\alpha}$. Thus, $\widehat{\Sigma}_R$ entries are determined by the Lagrangian multipliers of primal program. More specifically, we have
\begin{equation}
(\widehat{\Sigma}_R)_{ij}= \left\{ \begin{array}{lcl}
		0 & \operatorname{for} & i = j \\
		0 & \operatorname{for} & i \neq j, |\bigl( \widehat{J}_M \bigr)_{ij}| < \lambda \\
		\widehat{\alpha}_{ij} \sign \bigl( \bigl( \widehat{J}_M \bigr)_{ij} \bigr) & \operatorname{for} & i \neq j, |\bigl( \widehat{J}_M \bigr)_{ij}| = \lambda,
		\end{array}\right.
\label{SigmaR_JM_relation_sample_version}
\end{equation}
where $\widehat{\alpha}_{ij} \geq 0$ are the Lagrangian multipliers of primal program \eqref{convex_prog_primal_sample_case}. \\

\bprf
We prove Lemma \ref{optimal_solution_characterization_sample_version} here and Lemma \ref{optimal_solution_characterization} is a special case of that when $\gamma$ is set to zero and $\widehat{\Sigma}^n$ is substituted with $\Sigma^*$.

For any $\lambda > 0$ and $\gamma \geq 0$, the optimization problem \eqref{convex_prog_primal_sample_case} is a convex programming where the objective function is strictly convex. Therefore, if the minimum is achieved it is unique. Since off-diagonal entries of $J_M$ are bounded according to constraints, the only issue for minimum achievement may arises for unbounded diagonal entries. It is shown in \cite{Ravikumar&etal:08Arxiv} that if diagonal entries of $\widehat{\Sigma}^n$ are strictly positive, the function is coercive with respect to diagonal entries and therefore here is no issue regarding unbounded diagonal entries. Thus, the minimum is attained in $J_M \succeq 0$. But since when $J_M$ approaches the boundary of positive definite cone, the objective function goes to infinity, the solution is attained in the interior of the cone $J_M \succ 0$. After showing that the unique minimum is achieved, let us characterize the minimum. \\

Considering $\alpha_{ij}$ as Lagrangian multipliers of inequality constraints of program \eqref{convex_prog_primal_sample_case}, the Lagrangian function is
\begin{equation}
\mathcal{L} (J_M, \alpha) = \langle \widehat{\Sigma}^n,J_M \rangle - \log \det J_M + \gamma \|J_M\|_{1, \operatorname{off}} + \sum_{i \neq j} \alpha_{ij} \bigl[ \bigl| \bigl( J_M \bigr)_{ij} \bigr| - \lambda \bigr] .
\end{equation}
We skipped positive definiteness constraint in writing Lagrangian function since it is inactive. Based on standard convex optimization theory, the matrix $\widehat{J}_M \succ 0$ is the optimal solution if and only if it satisfies KKT conditions. It should minimize the Lagrangian which happens if and only if $0$ belongs to the subdifferential of Lagrangian or equivalently there exists a matrix $\widehat{Z}$ such that
\begin{equation}
\widehat{\Sigma}^n - \widehat{J}_M^{-1} + \widehat{Z} = 0,
\end{equation}
where $\widehat{Z} = \widehat{Z}_{\alpha} + \gamma \widehat{Z}_{\gamma}$. Matrix $\widehat{Z}_{\gamma} \in \partial \|\widehat{J}_M\|_{1, \operatorname{off}}$ and $\widehat{Z}_{\alpha}$ is
\begin{equation}
(\widehat{Z}_{\alpha})_{ij}= \left\{ \begin{array}{lcl}
		0 & \operatorname{for} & i=j \\
		\in \widehat{\alpha}_{ij} . [-1,1] & \operatorname{for} & i\neq j, \bigl( \widehat{J}_M \bigr)_{ij}=0 \\
		\widehat{\alpha}_{ij} \sign \bigl( \bigl( \widehat{J}_M)_{ij} \bigr) \bigr) & \operatorname{for} & i\neq j, \bigl( \widehat{J}_M \bigr)_{ij} \neq 0,
		\end{array}\right.
\end{equation}
for some Lagrangian multipliers $\widehat{\alpha}_{ij} \geq 0$. The solution should also satisfy complementary slackness conditions $\widehat{\alpha}_{ij}.\bigl[ \big| \bigl( \widehat{J}_M \bigr)_{ij} \bigr| - \lambda \bigr] = 0$ for $i \neq j$. Applying this condition to above $\widehat{Z}_{\alpha}$ representation, results to \eqref{Z_matrix_original} form proposed in the lemma.
\eprf

\section{Proof of Theorem~\ref{thm:decomp}}\label{proof:decomp}

First note that as mentioned in the remark in section \ref{sec:  algo. formulation}, the pair $\bigl( \widehat{J}_M,\widehat{\Sigma}_R \bigr)$ given by optimization program gives a decomposition $\Sigma^* = \widehat{J}_M^{-1} - \widehat{\Sigma}_R$ which is desired.

Next, in order to prove the equivalence, we show that there is a one to one correspondence between the specified conditions (A.0)-(A.3) for valid decomposition and the characterization of optimal solution of optimization program given in lemma \ref{optimal_solution_characterization}. We go through each of these conditions one by one in the following lines. \\
Condition (A.0) is considered in optimization program as positive definiteness of Markov matrix $J_M$. \\
Condition (A.1) is exactly the primal constraint $\|J_M^*\|_{\infty, \operatorname{off}} \leq \lambda$. \\
Condition (A.2) is exactly the same as relation \eqref{SigmaR_JM_relation} where diagonal entries of residual covariance matrix are zero and its off-diagonal entries can be nonzero only if the absolute value of corresponding entry in Markov matrix takes the maximum value $\lambda$. \\
Condition (A.3) is exactly the same as inequality $\widehat{\alpha}_{ij} \geq 0$. \\
In the above lines, we covered one by one correspondence for conditions (A.0)-(A.3). But note that we also covered all the equalities and inequalities that characterize unique optimal solution of optimization program. In other words by above correspondence we proved that both of the following derivations are true where second one is the reverse of first one. On one hand, any optimal solution of optimization program gives a valid decomposition under desired conditions. On the other hand, any valid decomposition under desired conditions is a solution of proposed optimization program. Thus, we can infer that these two are exactly equivalent and the result is proved.  Since the solution of optimization program is unique and according to the equivalence between this solution and decomposition under those conditions,   uniqueness is also established.
\qed

\section{Proof of Theorem \ref{sample case theorem}} \label{appendix: proof of sample case theorem}

In this appendix, we first mention an outline of the primal-dual witness method and then provide the detailed proof of the theorem.

\subsection{Primal-Dual Witness Method}
First, continuing the proof outline presented in section \ref{sec:proof outline}, we provide an outline of the primal-dual witness method steps in order to establish equivalence between optimal solutions of the original \eqref{convex_prog_primal_sample_case} and the modified \eqref{convex_prog_primal_sample_case_modified} optimization programs.

\begin{enumerate}

\item The primal witness matrix $\widetilde{J}_M$ is defined as in \eqref{convex_prog_primal_sample_case_modified}.

\item The dual witness matrix is set as $\widetilde{Z} = - \widehat{\Sigma}^n + \widetilde{J}_M^{-1}$. It is defined in this way to satisfy original program optimal solution characterization mentioned in appendix \ref{appendix: optimal solution characterization}.

\item We need to check the following feasibility conditions under which the modified program solution is equivalent to the solution of original one:
\begin{enumerate}
\item $\|\widetilde{J}_M\|_{\infty , \operatorname{off}, S} \leq \lambda$: Since we relaxed the $\ell_\infty$ bounds on off-diagonal entries in set $S$, we need to make sure that the modified solution satisfies this bound in order to have equivalence between modified and original programs solutions.
\item Set $\bigl( \widetilde{Z}_\alpha \bigr)_{S_R} = \bigl( - \widehat{\Sigma}^n + \widetilde{J}_M^{-1} - \gamma \bigl( \widetilde{Z}_\gamma \bigr) \bigr)_{S_R}$ where $\widetilde{Z}_{\gamma} \in \partial \|\widetilde{J}_M\|_{1, \operatorname{off}}$. Note that since $|\bigl( \widetilde{J}_M \bigr)_{ij}| = \lambda \neq 0$ for any $(i,j) \in S_R$, then $\widetilde{Z}_{\gamma}$ and therefore above equation is well-defined. Now we need to check: $\bigl( \widetilde{Z}_\alpha \bigr)_{ij} \bigl( \widetilde{J}_M \bigr)_{ij} \geq 0$ for all $(i,j) \in S_R$. This means that they have the same sign or one of them is zero. We need this condition for equivalence between solutions because Lagrangian multipliers in original program \eqref{convex_prog_primal_sample_case} corresponding to inequality constraints should be nonnegative.
\item $\|\widetilde{Z}\|_{\infty, S_M^c} < \gamma$: According to the $\bigl( J_M \bigr)_{S_M^c} = 0$ constraint in the modified program, all the inequality constraints become inactive in the original one when desired $\widehat{J}_M = \widetilde{J}_M$ equality is satisfied. Then, complementary slackness condition enforce all the Lagrangian multipliers corresponding to set $S_M^c$ to be zero. These can be satisfied by the above strict dual feasibility. Also note that having zero Lagrangian multipliers results in zero residual entries, i.e., $\bigl( \widetilde{\Sigma}_R \bigr)_{S_M^c} = 0$  and therefore $\| \widetilde{\Delta}_R \|_{\infty,S_M^c}=0$ when this feasibility condition is satisfied.
\end{enumerate}

\end{enumerate}

Also note that we dropped the positive-definiteness constraint $\Sigma_M - \Sigma_R \succ 0$ in the proof outline. Thus, in addition to above conditions, we also need to show that $\widetilde{\Sigma} = \widetilde{\Sigma}_M - \widetilde{\Sigma}_R \succ 0$ in the modified program.


Before we state the detailed proof for the theorem, we introduce a pair of definitions which are used in the analysis. Let us define matrix $E$ as difference between sample covariance matrix and the exact covariance matrix
\begin{equation}
\label{sample cov matrix error}
E := \widehat{\Sigma}^n - \Sigma^*.
\end{equation}
We also define $R \bigl( \widetilde{\Delta}_J \bigr)$ as the difference between $\widetilde{J}_M^{-1}$ and its first order Taylor expansion around $J_M^*$. Recall that $\widetilde{\Delta}_J$ was defined as $\widetilde{\Delta}_J := \widetilde{J}_M - J_M^*$. According to results for first order derivative of inverse function $J_M^{-1}$ \cite{Boyd:book}, the remainder is
\begin{equation} \label{remainder definition}
R\bigl( \widetilde{\Delta}_J \bigr) = \widetilde{J}_M^{-1} - {J_M^*}^{-1} + {J_M^*}^{-1} \widetilde{\Delta}_J {J_M^*}^{-1}.
\end{equation}


\subsection{Proof of the Theorem}
Exploiting lemmata mentioned in Appendix \ref{appendix: aux lemmas}, the Theorem \ref{sample case theorem} is proved as follows: \\
\noindent \bprf
According to the sample error bound mentioned in Lemma \ref{sample error bound}, we have $\| E \|_\infty \leq \overline{\delta}_f(p^\tau;n)$ for some $\tau>2$ with probability greater than or equal to $1-1/{p^{\tau-2}}$. In the discussion after this, it is assumed that the above bound for $\|E\|_\infty$ is satisfied and therefore the following results are valid with probability greater than or equal to $1-1/{p^{\tau-2}}$.

By choosing $\gamma = \frac{m}{\alpha} \overline{\delta}_f(p^\tau;n)$, we have $\| E \|_\infty \leq \frac{\alpha}{m} \gamma$ as desired for Lemma \ref{Feasibility Lemma}. Choosing $\lambda_\delta$ as in \eqref{lambda_delta} (compatible with what mentioned in the theorem), we only need to show that the other bound on $\| R \|_\infty$ is also satisfied to be able to apply Lemma \ref{Feasibility Lemma}. 
As stated in the remark after Theorem \ref{sample case theorem}, the bound on sample complexity is not asymptotic and we assume the following lower bound on the number of samples which is compatible with the asymptotic form mentioned in the theorem:
\begin{equation} \label{exact sample complexity}
n > \overline{n}_f \Biggl( p^\tau ; 1 / \max \biggl \{ v_*, 4 l d \Bigl( 1+\frac{m}{\alpha} \Bigr) K_{SS} K_M \max \Bigl \{ 1 , \frac{4}{l-1} \Bigl( 1+\frac{m}{\alpha} \Bigr) K_{SS} K_M^2 \Bigr\} \biggr \} \Biggr ),
\end{equation}
for some $l>1$. Because of monotonic behaviour of the tail function, for any $n$ satisfying above bound, we have:
\begin{equation} \label{delta_f bound}
\overline{\delta}_f \bigl( p^\tau;n) \leq \min \biggl \{ \frac{1}{v_*} , \frac{1}{4 l d (1+\frac{m}{\alpha}) K_{SS} K_M} , \frac{l-1}{16 l d (1+\frac{m}{\alpha})^2 K_{SS}^2  K_M^3 } \biggr \},
\end{equation}
According to the selection for regularization parameters $\lambda_\delta$ and $\gamma$ and the bound on sample error $\|E\|_\infty$, we have:
\begin{align}
r :=  2 K_{S S_R} \lambda_\delta + 2 K_{S S} \bigl( \|E\|_\infty + \gamma \bigr) & \leq \biggl[ \frac{ 4 K_{S S_R} K_{SS}}{1 - 2 K_{S S_R}} \Bigl( 1 + \frac{\alpha}{m} \Bigr) \frac{m}{\alpha} + 2 K_{S S} \Bigl( 1 + \frac{m}{\alpha} \Bigr) \biggr] \overline{\delta}_f (p^\tau;n) \\
& = 2 K_{S S} \Bigl( 1 + \frac{m}{\alpha} \Bigr) \overline{\delta}_f (p^\tau;n) \frac{1}{1- 2 K_{S S_R}} \nonumber \quad (=\lambda_\delta) \\
& < 4 K_{S S} \Bigl( 1 + \frac{m}{\alpha} \Bigr) \overline{\delta}_f (p^\tau;n), \nonumber
\end{align}
where in the last inequality, we used the second condition is assumption (A.4) that $K_{S S_R} < 1/4$. Note that second line is equal to $\lambda_\delta$ since we assigned the same value in \eqref{lambda_delta}. Applying the bound \eqref{delta_f bound} on above inequality, we have
\begin{align}
2 K_{S S_R} \lambda_\delta + 2 K_{S S} \bigl( \|E\|_\infty + \gamma \bigr) & < \min \left\{ \frac{1}{l d K_M} , \frac{l-1}{4 l d (1+\frac{m}{\alpha}) K_{S S} K_M^3} \right \} \\
& \leq \min \left\{ \frac{1}{l d K_M} , \frac{l-1}{2 l d K_{S S} K_M^3} \right \}. \nonumber
\end{align}
Thus, the conditions for Lemma \ref{control delta} are satisfied and we have
\begin{equation}
\label{multiple bounds}
\| \widetilde{\Delta}_J \|_{\infty,S} \leq 2 K_{S S_R} \lambda_\delta + 2 K_{S S} \bigl( \|E\|_\infty + \gamma \bigr) \leq \lambda_\delta < 4 K_{S S} \Bigl( 1 + \frac{m}{\alpha} \Bigr) \overline{\delta}_f (p^\tau;n).
\end{equation}
Above inequalities tell us multiple things. First, since the error $\| \widetilde{\Delta}_J \|_{\infty,S}$ is bounded by $\lambda_\delta$, the $\widetilde{J}_M$ entries in set $S$ can not deviate from exact one $J_M^*$ more than $\lambda_\delta$. We also assumed that the off-diagonal entries in $J_M^*$ are bounded by $\lambda^*$. Therefore according to the definition of $\lambda_\delta := \lambda - \lambda^*$, the entries in $\bigl( \widetilde{J}_M \bigr)_{\operatorname{off},S}$ are bounded by $\lambda$ and therefore the condition (a) for feasibility of primal-dual witness method is satisfied, i.e., we have $\|\widetilde{J}_M\|_{\infty , \operatorname{off}, S} \leq \lambda$. Second, since $\| \widetilde{\Delta}_J \|_{\infty,S_R} = \lambda_\delta$, we have $\| \widetilde{\Delta}_J \|_{\infty,S} \leq \| \widetilde{\Delta}_J \|_{\infty,S_R}$ and therefore $\| \widetilde{\Delta}_J \|_{\infty} = \| \widetilde{\Delta}_J \|_{\infty,S_R} = \lambda_\delta$ which results the following error bound
\beq \| \widetilde{\Delta}_J \|_{\infty} := \| \widetilde{J}_M - J_M^* \|_{\infty} \leq 4 K_{S S} \Bigl( 1 + \frac{m}{\alpha} \Bigr) \overline{\delta}_f (p^\tau;n). \label{J_M_error non-asymptotic} \eeq 
Furthermore, $\| \widetilde{\Delta}_J \|_{\infty} < \frac{1}{l d K_M }$ bound can be concluded from above inequality by substituting $\overline{\delta}_f (p^\tau;n)$ from \eqref{delta_f bound}. Thus, the condition for Lemma \ref{Conrol remainder} is satisfied and we have the following bound on the remainder term
\begin{align}
\| R \bigl( \widetilde{\Delta}_J \bigr) \|_\infty & \leq \frac{l}{l-1} d \| \widetilde{\Delta}_J \|_\infty^2 K_M^3 \\
& \leq \frac{16 l}{l-1} d K_M^3 K_{S S}^2 \Bigl( 1 + \frac{m}{\alpha} \Bigr)^2 \bigl[ \overline{\delta}_f (p^\tau;n) \bigr]^2 \nonumber \\
& = \biggl[\frac{16 l}{l-1} d K_M^3 K_{S S}^2 \Bigl( 1 + \frac{m}{\alpha} \Bigr)^2 \overline{\delta}_f (p^\tau;n) \biggr] \overline{\delta}_f (p^\tau;n) \nonumber \\
& \leq \overline{\delta}_f (p^\tau;n) = \frac{\alpha}{m} \gamma,
\end{align}
where in the second inequality, we used error bound in \eqref{J_M_error non-asymptotic} and the last inequality is concluded from bound \eqref{delta_f bound}.

Now the conditions for Lemma \ref{Feasibility Lemma} are satisfied and therefore we have the upper bound on $\| \widetilde{\Delta}_R \|_{\infty, S_R} < C_3 \gamma$ and the strict dual feasibility on $S_M^c$. Second result satisfies condition (c) of the primal-dual witness method feasibility conditions. The upper bound on $\| \widetilde{\Delta}_R \|_{\infty, S_R}$ in conjunction with the lower bound on $\bigl( \Sigma_R^* \bigr)_{\operatorname{min}} > C_3 \gamma$ (mentioned in the theorem), ensures that the sign of $\Sigma_R^*$ and $\widetilde{\Sigma}_R$ are the same which results that the condition (b) of the feasibility conditions for primal-dual witness method is satisfied. Since all three conditions (a)-(c) are satisfied, we have equivalence between the modified program and the original one under conditions specified in the theorem. It gives us both results (a) and (b) in the theorem. Then by assuming lower bound on minimum nonzero value of $J_M^*$, the result in part (c) is also proved.

As mentioned before, we need to show that the dropped constraint $\Sigma = \Sigma_M - \Sigma_R \succ 0$ is also satisfied.
Since the conditions for Corollary \ref{spectral norm Sigma} in Appendix \ref{appendix:spectral norm Sigma} are satisfied, we have the spectral norm error bound \eqref{Sigma_error non-asymptotic} on overall covariance matrix $\Sigma$.
Applying the inverse tail function for Gaussian distribution in \eqref{inverse tail function} to assumption (A.6) results that the minimum eigenvalue of exact covariance matrix $\Sigma^*$ satisfies lower bound $\lambda_{\min} (\Sigma^*) \geq \bigl( C_4 + \frac{m}{\alpha} C_3 \bigr) d \overline{\delta}_f (p^{\tau};n) + C_5 d^2 \bigl[ \overline{\delta}_f (p^{\tau};n) \bigr]^2$ where $C_6 := \bigl( C_4 + \frac{m}{\alpha} C_3 \bigr) \sqrt{2 q^2}$ and $C_7 := 2q^2 C_5$. Then by exploiting Weyl's theorem (Theorem 4.3.1 in \cite{Horn&Johnson:book}), the estimated covariance matrix $\widehat{\Sigma}$ is positive definite and thus valid. Therefore, the result is proved.
\eprf

\section{Auxiliary Lemmas} \label{appendix: aux lemmas}

First, the tail condition for a probability distribution is defined as follows. \\
\noindent \begin{definition}[Tail Condition]
The random vector $X$ satisfies tail condition with parameters $f$ and $v_*$ if there exists a constant $v_* \in (0,\infty)$ and function $f: \mathbb{N} \times (0,\infty) \rightarrow (0,\infty)$ such that for any $(i,j) \in V \times V$:
\begin{equation}
\mathbb{P} [|\widehat{\Sigma}^n - \Sigma^*_{ij}| \geq \delta ] \leq \frac{1}{f(n,\delta)} \ \operatorname{for \ all} \delta \in (0,\frac{1}{v_*}].
\end{equation}
\end{definition}
Note that since the function $f(n,\delta)$ is an increasing function of both variables $n$ and $\delta$, we define the inverse functions $\overline{n}_f(r;\delta)$ and $\overline{\delta}_f(r;n)$ with respect to variables $n$ and $\delta$ respectively (when the other argument is fixed), where $f(n,\delta)=r$.

\subsection{Concentration Bounds}

From Lemma 1 in \cite{Ravikumar&etal:08Arxiv}, we have the following concentration bound for the empirical covariance matrix of Gaussian random variables.
\begin{lem}[\cite{Ravikumar&etal:08Arxiv}]
\label{Gaussian concentration bound}
Consider a set of Gaussian random variables with covariance matrix $\Sigma^*$. Given n i.i.d. samples, the sample covariance matrix $\widehat{\Sigma}^n$ satisfies
\begin{equation}
\mathbb{P} [|\widehat{\Sigma}^n_{ij} - \Sigma^*_{ij}| > \delta] \leq 4 \exp \left\{-\frac{n \delta^2}{2 q^2} \right\} \ \operatorname{for \ all} \ \delta \in (0,q),
\end{equation}
for some constant $q > 0$.
\end{lem}

Thus the tail function for Gaussian random vector takes the exponential form with the following corresponding inverse functions:

\begin{equation}
\overline{n}_f (r;\delta) = \frac{2 q^2 \log(4r)}{\delta^2}, \quad \overline{\delta}_f (r;n) = \sqrt{ \frac{2 q^2 \log(4r)}{n} } \label{inverse tail function}
\end{equation}

Applying above Lemma, we get the following bound for sampling error.

\begin{lem}[\cite{Ravikumar&etal:08Arxiv}] \label{sample error bound}
For any $\tau>2$ and sample size $n$ such that $\overline{\delta}_f (p^{\tau};n) < 1/v_*$, we have
\beq \mathbb{P} \bigl[ \| E \|_\infty \geq \overline{\delta}_f (p^{\tau};n) \bigr] \leq \frac{1}{p^{\tau-2}} \rightarrow 0. \eeq
\end{lem}

\subsection{Feasibility Conditions}
In the following lemma, we propose some conditions to bound the residual error $\| \widetilde{\Delta}_R \|_{\infty, S_R}$ and also satisfy the condition (c) of feasibility conditions required for equivalence between the witness solution and the original one.
\begin{lem} \label{Feasibility Lemma}
Suppose that
\begin{align}
& \max \left\{ \| R \|_\infty, \| E \|_\infty \right\} \leq \frac{\alpha}{m} \gamma , \\
& \lambda_\delta = \frac{2 K_{SS}}{1 - 2 K_{S S_R}} \Bigl( 1 + \frac{\alpha}{m} \Bigr) \gamma , \label{lambda_delta}
\end{align}
then

a) $\| \widetilde{\Delta}_R \|_{\infty, S_R} \leq C_3 \gamma$ for some $C_3 > 0$.

b) $\| \widetilde{Z} \|_{\infty, S_M^c} < \gamma$.
\end{lem}

\bprf
Applying definitions \eqref{sample cov matrix error} and \eqref{remainder definition} to optimality condition considered in second step of primal-dual witness method construction, gives the following equivalent equation
\begin{equation}
\label{optimal solution_characterization_modified_version}
{J_M^*}^{-1} \widetilde{\Delta}_J {J_M^*}^{-1} - \Sigma_R^* - R\bigl( \widetilde{\Delta}_J \bigr) + E + \widetilde{Z} = 0.
\end{equation}
Above equation is a $p \times p$ matrix equation. We can rewrite it as a linear equation with size $p^2$ if we use the vectorized form of matrices. Vectorized form of a matrix $D \in \mathbb{R}^{p \times p}$ is a column vector $\overline{D} \in \mathbb{R}^{p^2}$ which is composed by concatenating the rows of matrix $D$ in a single column vector. In the vectorized form, we have
\begin{equation}
\operatorname{vec} \bigl( {J_M^*}^{-1} \widetilde{\Delta}_J {J_M^*}^{-1} \bigr)  = \bigl( {J_M^*}^{-1} \otimes {J_M^*}^{-1} \bigr)  \overline{\widetilde{\Delta}}_J = \Gamma^* \overline{\widetilde{\Delta}}_J.
\end{equation}
Decomposing the vectorized form of \eqref{optimal solution_characterization_modified_version} into three disjoint partitions $S$, $S_R$ and $S_M^c$ gives the following decomposed form
\begin{equation} \label{optimal solution_characterization_modified_version_decomposed}
\left[ \begin{array}{ccc}
\Gamma^*_{S S} & \Gamma^*_{S S_R} &\Gamma^*_{S S_M^c} \\
\Gamma^*_{S_R S} & \Gamma^*_{S_R S_R} &\Gamma^*_{S_R S_M^c} \\
\Gamma^*_{S_M^c S} & \Gamma^*_{S_M^c S_R} &\Gamma^*_{S_M^c S_M^c}
\end{array} \right]
\left[ \begin{array}{c}
\Bigl( \overline{\widetilde{\Delta}}_J \Bigr) _S \\
\overrightarrow{\lambda_\delta} \\
0
\end{array} \right]
-
\left[ \begin{array}{c}
0 \\
\bigl( \overline{\Sigma}_R^* \bigr) _{S_R} \\
0
\end{array} \right]
+
\left[ \begin{array}{l}
\bigl( -\overline{R} + \overline{E} + \overline{\widetilde{Z}} \bigr) _S \\
\bigl( -\overline{R} + \overline{E} + \overline{\widetilde{Z}} \bigr) _{S_R} \\
\bigl( -\overline{R} + \overline{E} + \overline{\widetilde{Z}} \bigr) _{S_M^c}
\end{array} \right]
= 0,
\end{equation}
where we used the equalities $\bigl( \overline{\widetilde{\Delta}}_J \bigr) _{S_R} = \overrightarrow{\lambda_\delta}$ and $\bigl( \overline{\widetilde{\Delta}}_J \bigr) _{S_M^c} = 0$. Note that vector $\overrightarrow{\lambda_\delta}$ only includes $\pm \lambda_\delta$ entries according to the constraints in the modified program. Also note that $\Sigma_R^*$ is zero in sets $S$ and $S_M^c$. We also dropped the argument $\widetilde{\Delta}_J$ from remainder function $R \bigl( \widetilde{\Delta}_J \bigr) $ to simplify the notation.

Similar to the original program, the matrix $\widetilde{Z}$ is composed of two parts, $\widetilde{Z}_\beta$ and $\widetilde{Z}_\gamma$, i.e., $\widetilde{Z} = \widetilde{Z}_\beta + \gamma \widetilde{Z}_\gamma$. Matrix $\widetilde{Z}_\beta = \widetilde{\Sigma}_R$ from equation \eqref{SigmaR_JM_relation_sample_version_modified}, includes Lagrangian multipliers and $\widetilde{Z}_\gamma \in \partial \| \widetilde{J}_M \|_{1, \operatorname{off}}$. For set $S$, $\bigl( \widetilde{Z}_\beta \bigr)_S = 0$, since we don't have any constraint in the program and therefore the Lagrangian multipliers are zero. Applying this to the first row of equation \eqref{optimal solution_characterization_modified_version_decomposed} and since $\Gamma^*_{SS}$ is invertible, we have the following for error $\overline{\widetilde{\Delta}}_J$ in set $S$
\begin{equation} \label{DeltaJ error}
\bigl(  \overline{\widetilde{\Delta}}_J \bigr)_S = {\Gamma^*_{SS}}^{-1} \left[ - \Gamma_{S S_R}^* \overrightarrow{\lambda_\delta} + \overline{R}_S - \overline{E}_S - \gamma \bigl( \overline{\widetilde{Z}}_\gamma \bigr)_S \right],
\end{equation}
In set $S_R$, $\overline{\widetilde{Z}}_{S_R} = \bigl( \overline{\widetilde{\Sigma}}_R \bigr)_{S_R} + \gamma \bigl( \overline{\widetilde{Z}}_\gamma \bigr)_{S_R}$. Applying this to the second row of equation \eqref{optimal solution_characterization_modified_version_decomposed} results
\begin{equation}
\Gamma^*_{S_R S} \bigl(  \overline{\widetilde{\Delta}}_J \bigr)_S + \Gamma^*_{S_R S_R} \overrightarrow{\lambda_\delta} + \bigl(  \overline{\widetilde{\Delta}}_R \bigr)_{S_R} + \gamma \bigl( \overline{\widetilde{Z}}_\gamma \bigr)_{S_R} - \overline{R}_{S_R} + \overline{E}_{S_R} = 0,
\end{equation}
Recall that we defined $\widetilde{\Delta}_R := \widetilde{\Sigma}_R - \Sigma_R^*$. Substituting \eqref{DeltaJ error} in above equation results the following for error $\overline{\widetilde{\Delta}}_R$ in set $S_R$
\begin{align}
\bigl(  \overline{\widetilde{\Delta}}_R \bigr)_{S_R} = & -\Gamma^*_{S_R S} {\Gamma^*_{SS}}^{-1} \left[ - \Gamma_{S S_R}^* \overrightarrow{\lambda_\delta} + \overline{R}_S - \overline{E}_S - \gamma \bigl( \overline{\widetilde{Z}}_\gamma \bigr)_S \right] \\
& - \Gamma^*_{S_R S_R} \overrightarrow{\lambda_\delta} - \gamma \bigl( \overline{\widetilde{Z}}_\gamma \bigr)_{S_R} + \overline{R}_{S_R} - \overline{E}_{S_R}. \nonumber
\end{align}
Taking $\ell_\infty$ element-wise norm from above equation and using inequality $\| Ax \|_\infty \leq \gennorm{A}_\infty \| x \|_\infty$ for any matrix $A \in \mathbb{R}^{r \times s}$ and vector $x \in \mathbb{R}^s$, results the bound
\begin{align}
\| \widetilde{\Delta}_R \|_{\infty, S_R} \leq \gennorm{- \Gamma^*_{S_R S} {\Gamma^*_{SS}}^{-1} \Gamma_{S S_R}^* + \Gamma^*_{S_R S_R}}_\infty \lambda_\delta & + \gennorm{\Gamma^*_{S_R S} {\Gamma^*_{SS}}^{-1}}_\infty \left[ \| \overline{R}_S \|_\infty + \| \overline{E}_S \|_\infty + \gamma \right] \\
& +  \bigl( \| \overline{R}_{S_R} \|_\infty + \| \overline{E}_{S_R} \|_\infty + \gamma \bigr) , \nonumber
\end{align}
where we used the fact that $\| \overrightarrow{\lambda_\delta} \|_\infty = \lambda_\delta$ and $\| \widetilde{Z}_\gamma \|_\infty = 1$. Now if we apply the assumptions mentioned in the lemma,
\begin{align}
\label{Sigma_R_error non-asymptotic}
\| \widetilde{\Delta}_R \|_{\infty, S_R} \leq \left[ \frac{2 K_{SS} (m + \alpha)}{m(1 - 2 K_{S S_R})} \gennorm{- \Gamma^*_{S_R S} {\Gamma^*_{SS}}^{-1} \Gamma_{S S_R}^* + \Gamma^*_{S_R S_R}}_\infty \right. \nonumber \\
\left. + \bigl( 1 + \frac{2 \alpha}{m} \bigr) \bigl( 1 + \gennorm{\Gamma^*_{S_R S} {\Gamma^*_{SS}}^{-1}}_\infty \bigr) \right] \gamma = C_3 \gamma ,
\end{align}
which proves part (a) of the Lemma.

Now if we substitute \eqref{DeltaJ error} in the equation from third row of \eqref{optimal solution_characterization_modified_version_decomposed}, we have
\begin{equation}
\overline{\widetilde{Z}}_{S_M^c} = - \Gamma^*_{S_M^c S} {\Gamma^*_{SS}}^{-1} \left[ - \Gamma_{S S_R}^* \overrightarrow{\lambda_\delta} + \overline{R}_S - \overline{E}_S - \gamma \bigl( \overline{\widetilde{Z}}_\gamma \bigr)_S \right] - \Gamma^*_{S_M^c S_R} \overrightarrow{\lambda_\delta} + \overline{R}_{S_M^c} - \overline{E}_{S_M^c}.
\end{equation}
Taking $\ell_\infty$ element-wise norm from above equation gives the following bound
\begin{align}
\| \widetilde{Z} \|_{\infty, S_M^c} \leq \gennorm{\Gamma^*_{S_M^c S} {\Gamma^*_{SS}}^{-1} \Gamma_{S S_R}^* - \Gamma^*_{S_M^c S_R}}_\infty \lambda_\delta & + \gennorm{\Gamma^*_{S_M^c S} {\Gamma^*_{SS}}^{-1}}_\infty \left[ \| \overline{R}_S \|_\infty + \| \overline{E}_S \|_\infty + \gamma \right] \\
& + \| \overline{R}_{S_M^c} \|_\infty + \| \overline{E}_{S_M^c} \|_\infty , \nonumber
\end{align}
where we used the fact that $\| \widetilde{Z}_\gamma \|_\infty = 1$. Applying assumption (A.4) to above bound results
\begin{equation}
\| \widetilde{Z} \|_{\infty, S_M^c} \leq (1-\alpha) \lambda_\delta + (2-\alpha) \left[ \| R \|_\infty + \| E \|_\infty \right] + (1-\alpha) \gamma.
\end{equation}
Using assumptions stated in the Lemma, we have
\begin{align}
\| \widetilde{Z} \|_{\infty, S_M^c} & \leq \left[ \frac{2 K_{SS}}{1 - 2 K_{S S_R}} \Bigl( 1 + \frac{\alpha}{m} \Bigr) (1-\alpha) + (2-\alpha) \frac{2 \alpha}{m} + (1-\alpha) \right] \gamma \\
& < \left[ 4 K_{SS} \Bigl( 1 + \frac{\alpha}{m} \Bigr) (1-\alpha) + (2-\alpha) \frac{2 \alpha}{m} + (1-\alpha) \right] \gamma \nonumber \\
& < \left[ 4 K_{SS} \frac{m - (m-1) \alpha}{m} + \frac{4 \alpha}{m} + (1-\alpha) \right] \gamma \leq \gamma , \nonumber
\end{align}
where we used the bound on $K_{S S_R}$ in assumption (A.4) in the second inequality and the fact that $\alpha > 0$ in the third inequality. Final inequality is derived from assumption (A.5) which finishes the proof of part (b).
\eprf

\subsection{Control of Remainder}
In the following Lemma which is stated and proved in lemma 5 in \cite{Ravikumar&etal:08Arxiv}, the argument $\widetilde{\Delta}_J$ controls the remainder function behavior.
\begin{lem} \label{Conrol remainder}
Suppose that the element-wise $\ell_\infty$ bound $\| \widetilde{\Delta}_J \|_\infty \leq \frac{1}{l K_M d}$ for some $l>1$ holds. Then
\begin{equation}
R \bigl( \widetilde{\Delta}_J \bigr) = \bigl( {J^*_M}^{-1} \widetilde{\Delta}_J \bigr)^2 Q {J_M^*}^{-1},
\end{equation}
where $Q := \sum_{k=0}^\infty (-1)^k \bigl( {J^*_M}^{-1} \widetilde{\Delta}_J \bigr)^k$ with bound $\gennorm{Q^T}_\infty \leq \frac{l}{l-1}$. Also, in terms of element-wise $\ell_\infty$ norm, we have
\begin{equation}
\| R \bigl( \widetilde{\Delta}_J \bigr) \|_\infty \leq \frac{l}{l-1} d \| \widetilde{\Delta}_J \|_\infty^2 K_M^3.
\end{equation}
\end{lem}

\subsection{Control of $\widetilde{\Delta}_J$}
According to  the primal-dual witness solutions construction, we have the error bounds on $\widetilde{\Delta}_J$ within the sets $S_R$ and $S_M^c$ such that $\| \widetilde{\Delta}_J \|_{\infty, S_R} = \lambda_\delta$ and $\| \widetilde{\Delta}_J \|_{\infty, S_M^c} = 0$. In the following lemma, we propose some conditions to control the error $\| \widetilde{\Delta}_J \|_{\infty, S}$.

\begin{lem} \label{control delta}
Suppose that
\begin{equation}
r := 2 K_{S S_R} \lambda_\delta + 2 K_{S S} \bigl( \|E\|_\infty + \gamma \bigr) \leq \min \left\{ \frac{1}{l d K_M} , \frac{l-1}{2 l d K_{S S} K_M^3} \right \},
\end{equation}
then we have the following element-wise $\ell_\infty$ bound for $\bigl(\widetilde{\Delta}_J \bigr)_S$,
\begin{equation}
\|\widetilde{\Delta}_J \|_{\infty,S} \leq r.
\end{equation}
\end{lem}

The proof is within the same lines of Lemma 6 proof in \cite{Ravikumar&etal:08Arxiv} but with some modifications since the error $\| \widetilde{\Delta}_J \|_{\infty, S_R}$ is not zero and therefore the nonzero value $\lambda_\delta$ arises in the final result. Since the modified optimization program \eqref{convex_prog_primal_sample_case_modified} is different with the modified program in \cite{Ravikumar&etal:08Arxiv}, it is worth discussing about existing a unique solution for the modified optimization program \eqref{convex_prog_primal_sample_case_modified}. This uniqueness can be shown with similar discussion presented in Appendix \ref{appendix: optimal solution characterization} for uniqueness of the solution of original program \eqref{convex_prog_primal_sample_case}. We only need to show that there is no problem in uniqueness by removing the off-diagonal constaraints for set $S$ in the modified program. By Lagrangian duality, the $\ell_1$ penalty term $\gamma \|J_M\|_{1, \operatorname{off}}$ can be moved to constraints as $\|J_M\|_{1, \operatorname{off}} \leq C(\gamma)$ for some bounded $C(\gamma)$. Therefore, the off-diagonal entries in set $S$ where the corresponding constraints were relaxed in the modified program are still bounded because of this $\ell_1$ constraint. Hence, the modified program \eqref{convex_prog_primal_sample_case_modified} has a unique solution.

\subsection{Spectral norm error bound on overall covariance matrix $\Sigma = J_M^{-1}-\Sigma_R$} \label{appendix:spectral norm Sigma}
\begin{corollary} \label{spectral norm Sigma}
Under the same assumptions (excluding (A.6)) as Theorem \ref{sample case theorem}, with probability greater than $1-1/p^c$, the overall covariance matrix estimate $\widehat{\Sigma} = \widehat{\Sigma}_M - \widehat{\Sigma}_R$ satisfies spectral norm error bound
\beq \gennorm { \widehat{\Sigma} - \Sigma^* } \leq \Bigl( C_4 + \frac{m}{\alpha} C_3 \Bigr) d \overline{\delta}_f (p^{\tau};n) + C_5 d^2 \bigl[ \overline{\delta}_f (p^{\tau};n) \bigr]^2. \label{Sigma_error non-asymptotic} \
\eeq
\end{corollary}
\bprf
We first bound the spectral norm errors for the Markov and residual covariance matrices $\widehat{\Sigma}_M$ and $\widehat{\Sigma}_R$. Along the same lines as Corollary 4 proof in \cite{Ravikumar&etal:08Arxiv}, the spectral norm error $\gennorm { \widehat{\Sigma}_M - \Sigma_M^* }$ can be bounded as 
\beq \gennorm { \widehat{\Sigma}_M - \Sigma_M^* } \leq C_4 d \overline{\delta}_f (p^{\tau};n) + C_5 d^2 \bigl[ \overline{\delta}_f (p^{\tau};n) \bigr]^2, \eeq
where $C_4 = 4 \bigl( 1+\frac{m}{\alpha} \bigr) K_{SS} K_M^2$ and $C_5 =  \frac{16l}{l-1} \bigl( 1+\frac{m}{\alpha} \bigr)^2 K_{SS}^2 K_M^3$. \\
The spectral norm error $\gennorm { \widehat{\Sigma}_R - \Sigma_R^* }$ can be also bounded as
\beq \gennorm { \widehat{\Sigma}_R - \Sigma_R^* } \leq \gennorm {\widehat{\Sigma}_R - \Sigma_R^*}_{\infty} \leq d \| \widehat{\Sigma}_R - \Sigma_R^* \|_\infty \leq \frac{m}{\alpha} C_3 d \overline{\delta}_f (p^{\tau};n), \eeq
where the first inequality is the property of spectral norm which is bounded by $\ell_\infty$-operator norm, second inequality is a result of the fact that $\widehat{\Sigma}_R$ and $\Sigma_R^*$ has at most $d$ nonzero entries in each row (since $S_R \subset S_M$) and the last inequality is concluded from the upper bound on $\ell_\infty$ element-wise norm error on residual matrix estimation stated in part (a) of Theorem \ref{sample case theorem}. \\
Applying the above bounds to the overall covariance matrix estimation $\widehat{\Sigma} = \widehat{\Sigma}_M - \widehat{\Sigma}_R$ and using the triangular inequality for norms, the bound in \eqref{Sigma_error non-asymptotic} is proven.
\eprf

\section{Proof of Corollary \ref{Corollary: sparse cov est}} \label{Appendix:proof of sparse cov corollary}
\bprf
The result in this corollary is a special case of general result in Theorem \ref{sample case theorem} when $\lambda^*=0$ and some minor modifications are considered in problem formulation. Note that, it is expressed in assumption (A.1) that the off-diagonal entries of exact Markov matrix $J_M^*$ are upper bounded by some positive $\lambda^*$. In order to extend the proof to the case of $\lambda^*=0$ (The case in this corollary), we need some minor modifications. First, the identifiability assumptions (A.0)-(A.3) can be ignored and instead it is assumed that the Markov part $J_M^*$ (or equivalently $\Sigma_M^*$) is diagonal and the residual part $\Sigma_R^*$ has only nonzero off-diagonal entries. Since the diagonal Markov matrix and off-diagonal residual matrix do not have any nonzero overlapping entries, it is natural that we do not require any more identifiability assumptions. Then, with these new assumptions, the set $S_M$ is defined as $S_M := S_R \cup \{ (i,i) | i=1,...,p \}$ where $S_R$ is defined the same as \eqref{Sigma_R Support def} and also set $S$ is defined the same as \eqref{set S def} which results that set $S$ includes only diagonal entries. Thus, the off-diagonal entries belongs to sets $S_R$ and $S_M^c$. Since  $\Sigma_M^*$ is a diagonal matrix, all submatrices of $\Gamma^*$ which are indexed by sets $S_R$ or $S_M^c$ are complete zero matrices. The result is that the terms which are bounded in the mutual incoherence condition (A.4) are already zero and thus there is no need to consider those additional assumptions in the corollary. \\ By making these changes in the problem formulation, the result in Corollary \ref{Corollary: sparse cov est} can be proven within the same lines of general result proof in Theorem \ref{sample case theorem}. It is only required to change the constraint on set  $S_R$ in the modified optimization program to $\bigl( J_M \bigr)_{S_R} = \lambda \sign \Bigl( \bigl( \Sigma_R^* \bigr)_{S_R} \Bigr)$.
\eprf




\end{document}